\documentclass[10pt,twocolumn,letterpaper]{article}

\usepackage{cvpr}              
\usepackage[table]{xcolor}
\usepackage{tikz}
\definecolor{cvprblue}{rgb}{0.21,0.49,0.74}
\usepackage[pagebackref,breaklinks,colorlinks]{hyperref}
\usepackage{pifont}
\usepackage{multirow}

\def\paperID{*****} 
\def\confName{CVPR}
\def\confYear{2026}

\title{FlowDrive: Energy Flow Field for End-to-End Autonomous Driving}

\author{
    Hao Jiang\textsuperscript{1}, Zhipeng Zhang\textsuperscript{1},  Yu Gao\textsuperscript{2}, Zhigang Sun\textsuperscript{2},  Yiru Wang\textsuperscript{2}, Yuwen Heng\textsuperscript{2}, Shuo Wang\textsuperscript{2},\\ Jinhao Chai\textsuperscript{4}, Zhuo Chen\textsuperscript{1},
    Hao Zhao\textsuperscript{3}, Hao Sun\textsuperscript{2},
    Xi Zhang\textsuperscript{1},  Anqing Jiang\textsuperscript{2\textdagger}, Chuan Hu\textsuperscript{1\textdagger}\\
    \vspace{-10pt}
    \\
    \footnotesize
    \begin{tabular}{c}
        \textsuperscript{1}Shanghai Jiao Tong University
        \textsuperscript{2}Bosch Corporate Research, Shanghai, China
        \textsuperscript{3}AIR, Tsinghua University
        \textsuperscript{4}Shanghai University, Shanghai, China \\
        \vspace{0.2cm}
        \textdagger\,Corresponding author: \href{mailto:anqing.jiang@cn.bosch.com}{anqing.jiang@cn.bosch.com}, \href{mailto:chuan.hu@sjtu.edu.cn}{chuan.hu@sjtu.edu.cn}
    \end{tabular}
}

\begin{document}
\maketitle
\begin{abstract}

Recent advances in end-to-end autonomous driving leverage multi-view images to construct BEV representations for motion planning. In motion planning, autonomous vehicles need considering both hard constraints imposed by geometrically occupied obstacles (e.g., vehicles,  pedestrians) and soft, rule-based semantics with no explicit geometry (e.g., lane boundaries, traffic priors).  However, existing end-to-end frameworks typically rely on BEV features learned in an implicit manner, lacking explicit modeling of risk and guidance priors for safe and interpretable planning. To address this, we propose \textbf{FlowDrive}, a novel framework that introduces physically interpretable energy-based flow fields—including risk potential and lane attraction fields—to encode semantic priors and safety cues into the BEV space. These flow-aware features enable adaptive refinement of anchor trajectories and serve as interpretable guidance for trajectory generation. Moreover, FlowDrive decouples motion intent prediction from trajectory denoising via a conditional diffusion planner with feature-level gating, alleviating task interference and enhancing multimodal diversity. Experiments on the NAVSIM v2 benchmark demonstrate that FlowDrive achieves state-of-the-art performance with an EPDMS of 86.3, surpassing prior baselines in both safety and planning quality. The project is available at \href{https://astrixdrive.github.io/FlowDrive.github.io/.}{https://astrixdrive.github.io/FlowDrive.github.io/.}
\end{abstract}    
\section{Introduction}
\label{sec:intro}

\begin{figure}
  \centering
  \begin{subfigure}{1\linewidth}
      \centering
      \includegraphics[width=\linewidth]{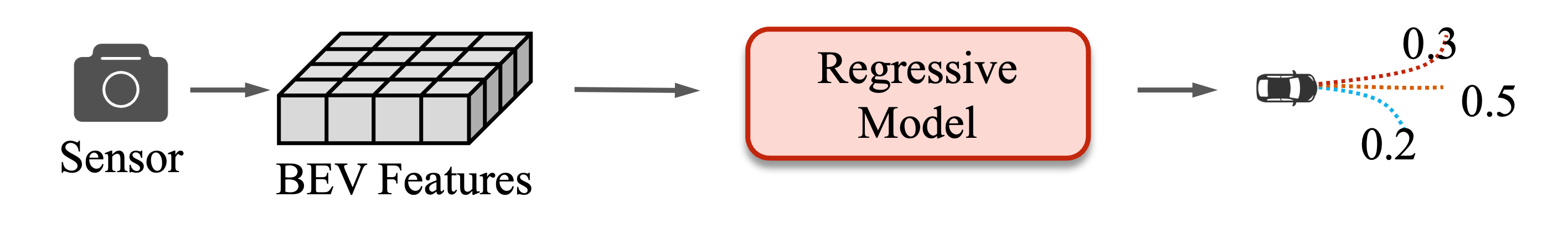}
      \caption{BEV-based Regression Paradigm}
      \label{fig:abstract_a}
      \vspace{-10pt}
  \end{subfigure}
  \vspace{0.3em}
  \begin{tikzpicture}
    \draw[dash pattern=on 6pt off 3pt, line width=1pt, color=black!30] (0,0) -- (8,0);
  \end{tikzpicture}

  \begin{subfigure}{1\linewidth}
      \centering
      \includegraphics[width=\linewidth]{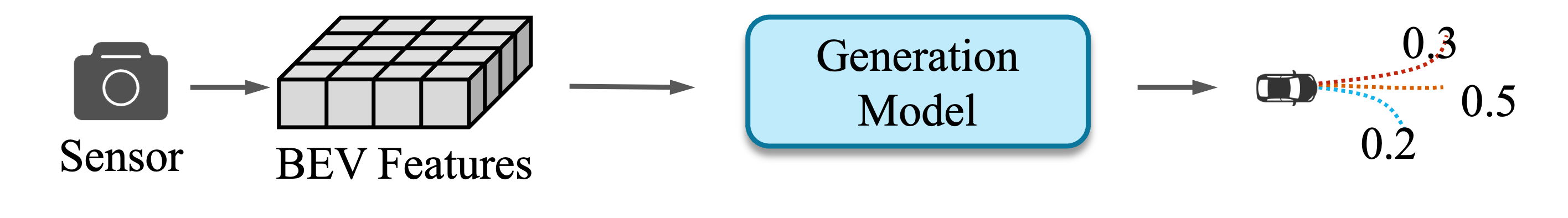}
      \caption{BEV-conditioned Generative Paradigm}
      \label{fig:abstract_b}
      \vspace{-10pt}
  \end{subfigure}
  \vspace{0.3em}
  \begin{tikzpicture}
    \draw[dash pattern=on 6pt off 3pt, line width=1pt, color=black!30] (0,0) -- (8,0);
  \end{tikzpicture}

  \begin{subfigure}{1\linewidth}
      \centering
      \includegraphics[width=\linewidth]{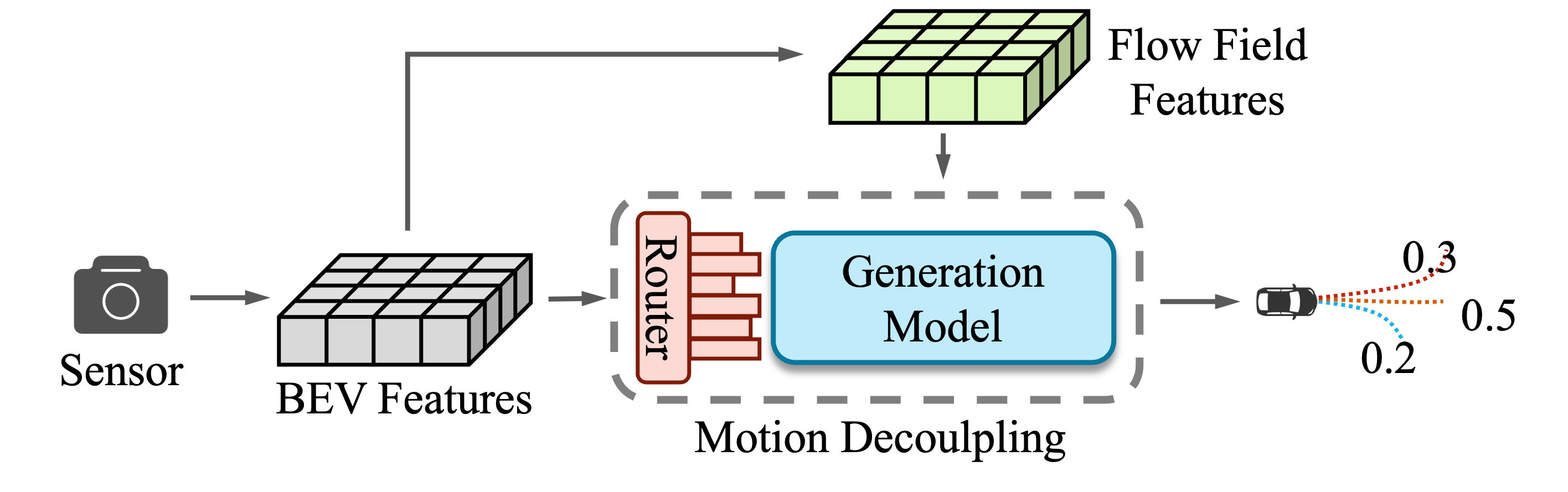}
      \caption{\textbf{Flow-guided Generative Paradigm (Ours)}} 
  \end{subfigure}
  \caption{\textbf{The comparison of different end-to-end paradigms.} \textbf{(a)} Regression-based paradigm that directly predicts trajectories from BEV representations. \textbf{(b)} Generative paradigm that samples trajectories conditioned on BEV features. \textbf{(c)} Our proposed decoupled generative paradigm based on BEV features and flow field features.}
  \label{fig:abstract}
\end{figure}
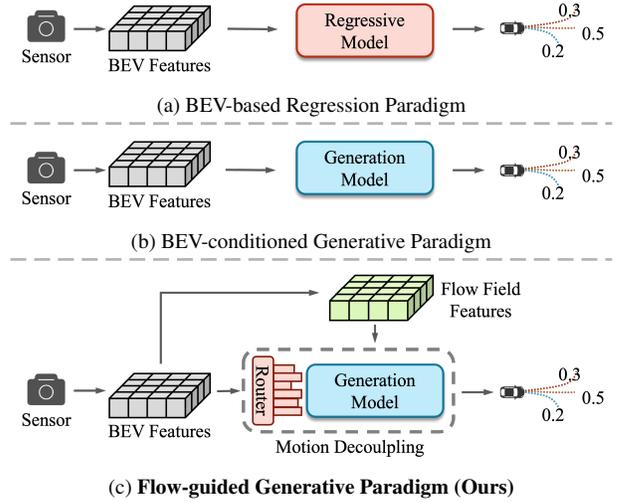

End-to-end autonomous driving has gained increasing attention in recent years due to its potential to simplify traditional modular pipelines and leverage large-scale data to jointly learn perception, prediction, and planning \cite{chen2024end,uniad,vad,vadv2,sparsedrive,transfuser,diffusiondrive,hydramdp++,wote,diffvla,diffsemanticfusion}. A prominent line of research focuses on using multi-view camera inputs to generate BEV representations, which serve as a spatially structured and compact format for downstream planning tasks \cite{uniad,vad,vadv2,transfuser,diffusiondrive,wote,ssr}. These BEV-based approaches have demonstrated promising performance in urban driving scenarios by providing a global understanding of the scene, including the layout of roads, surrounding vehicles, and other dynamic agents. 

While BEV representations offer a structured spatial view of the environment, they often lack explicit modeling of risk-sensitive semantics or rule-based planning priors, such as obstacle-induced repulsion or lane-centric guidance that are essential for safe and interpretable planning. Instead, these factors are typically embedded implicitly during end-to-end training, which hinders the model’s ability to reason about long-term safety and make high-level planning decisions with confidence. Furthermore, existing end-to-end planning frameworks often adopt a monolithic architecture that tightly couples high-level motion intent prediction with low-level trajectory generation \cite{uniad,vad,vadv2,sparsedrive,transfuser,diffusiondrive,hydramdp++,wote,ssr}. Such entanglement causes conflicting gradients and feature interference between motion intent prediction and trajectory generation. For example, high-level intent requires semantic abstraction, while precise trajectory generation demands fine-grained spatial accuracy that are hard to optimize jointly within a single feature space. This may limit task specialization and reduces generalization to diverse or long-tail scenarios.

To address these challenges, we propose FlowDrive, a novel end-to-end driving framework that introduces physically interpretable, energy-based guidance into the BEV space to  enhances BEV representations. Unlike prior methods that rely purely on implicit feature learning, FlowDrive introduces two learnable energy-based fields—risk potential fields and lane attraction fields—that represent obstacle avoidance and lane-level guidance through continuous spatial energy distributions. These fields represent spatial energy landscapes, where higher energy values correspond to regions of increased risk (\textit{e.g.}, obstacles, dynamic agents), and lower energy reflects goal-aligned or rule-compliant areas (\textit{e.g.}, lane centers). The gradients of these fields indicate the local direction of energy descent, offering structured, risk-aware guidance that enhances semantic alignment and spatial precision in trajectory planning beyond what purely learned features can provide. Building on these energy fields, we design a flow-aware anchor refinement module that adjusts initial trajectory anchors by aligning them with energy minima. This process improves trajectory initialization, enabling the planner to respond dynamically to scene geometry and driving semantics. To further integrate the structured guidance from flow fields into trajectory planning, we incorporate both flow features and BEV features as conditioning inputs into a diffusion-based planner. Unlike prior approaches that jointly optimize motion intent and trajectory prediction in a single decoder, we introduce a task-decoupled design that explicitly separates high-level intent reasoning from low-level trajectory denoising. This disentangled formulation enables each sub-task to focus on its distinct objective while leveraging shared structural context from the flow fields throughout the diffusion process. To summarize, the main contributions of this work are:


\begin{itemize}
    \item An energy-based flow field representation is proposed to explicitly to encode both geometry-induced constraints and rule-based semantics. These fields serve as structured and interpretable planning priors, guiding the planner away from risky regions and towards semantically meaningful, goal-directed areas.
    \item We introduce an flow-aware anchor refinement module that aligns coarse anchors with the flow field’s gradient structure, improving the spatial validity and intent consistency of generated trajectories.
    \item A task-decoupled diffusion planner separates intent prediction and trajectory generation, allowing targeted supervision and flow-conditioned decoding for diverse and goal-consistent trajectories.
\end{itemize}
\section{Related Works}
\label{sec:relate}

\subsection{End-to-End Autonomous Driving}

End-to-end autonomous driving has emerged as a promising paradigm that directly maps raw sensor inputs to planning decisions, bypassing traditional modular pipelines \cite{chen2024end,xiao2020multimodal}. Recent advances have introduced BEV-based frameworks, which transform sensor inputs into BEV representations to facilitate downstream trajectory planning \cite{transfuser,vad,vadv2,diffusiondrive,diffvla,hydramdp,hydramdp++}. These approaches are commonly categorized into regression-based (Fig.~\ref{fig:abstract_a}) and generation-based methods (Fig.~\ref{fig:abstract_b}), depending on the underlying mechanism of trajectory generation. Regression-based methods predict future waypoints or control signals by directly regressing from BEV features to fixed-format trajectories. For example, UniAD \cite{uniad} utilizes learnable task queries to extract BEV features for multitask learning, unifying detection, prediction, and planning in an unified framework. VAD \cite{vad} compresses scene semantics into compact vectorized representations to improve planning efficiency, while VADv2 \cite{vadv2} shifts toward multi-modal planning by selecting from a large discrete anchor set using scoring and sampling. Hydra-MDP \cite{hydramdp} and Hydra-MDP++ \cite{hydramdp++} refine this paradigm by introducing trajectory clustering and rule-based scorers to balance efficiency with diversity and safety.
Generation-based methods, by contrast, aim to model the distribution of future trajectories using generative frameworks such as VAEs \cite{genad}, diffusion models \cite{diffusiondrive,diffvla,transdiffuser}, or flow matching \cite{goalflow,moflow}. These methods produce diverse trajectory samples conditioned on scene context and driving intent. GenAD \cite{genad} adopts a VAE that encodes trajectory distributions via latent variables and uses a GRU decoder for path generation. Some diffusion-based methods \cite{diffusiondrive, diffvla, transdiffuser} apply the denoising diffusion process to anchors or real trajectories for high-quality generation. GoalFlow \cite{goalflow} formulates trajectory generation as flow matching over goal points, preserving structure and intention.
\subsection{Flow Field Representations in Planning}

Flow field-based representations have been widely explored in robotics \cite{potential,gradient,singularity,motion} and crowd simulation \cite{directing,yang2020review} as tools to model spatial influence, navigation guidance, and dynamic interactions. Early works have used potential fields and vector fields to represent obstacle repulsion and goal attraction forces \cite{hwang1992potential,warren1989global,chen2016uav}, serving as interpretable and reactive guidance signals.
In autonomous driving, flow fields have been primarily applied in rule-based path planning algorithms \cite{li2021optimization,ji2023tripfield,gao2024trajectory}. Despite their potential, flow field representations have not been effectively integrated into end-to-end autonomous driving frameworks. In this work, we bridge this gap by embedding energy-based flow fields into the BEV space as dense, physically interpretable priors that provide structured, safety-aware guidance and enhance the planner’s understanding of scene topology, dynamic risk, and behavioral intent.

\subsection{Multi-Task Learning}

Multi-task learning (MTL) has been widely adopted across computer vision, natural language processing, and recommendation systems to exploit shared inductive biases and reduce overfitting through parameter sharing \cite{zhang2018overview,tang2020progressive,zhang2021survey,crawshaw2020multi,vandenhende2021multi}. MoE \cite{moe} and MMoE \cite{mmoe} leverage gating over expert pools to enable conditional computation and task-aware sharing.
SNR \cite{snr} uses task-conditioned sparse routing to reduce cross-task interference. PLE \cite{ple} stacks shared and task-specific experts with gated cross-task fusion to enable controlled information sharing and disentangled task features. In our work, unlike methods that jointly optimize on the final captured features \cite{jiang2024hybrid,artemis}, we integrate multi-stage decoder features with gated selective sharing to decouple motion-mode prediction from trajectory generation, leveraging cross-layer semantics and fine-grained spatial cues to reduce gradient interference and improve intent accuracy and trajectory feasibility.
\section{Method}
\label{sec:method}

\begin{figure*}[t!]
    \centering
    \includegraphics[width=1.\linewidth]{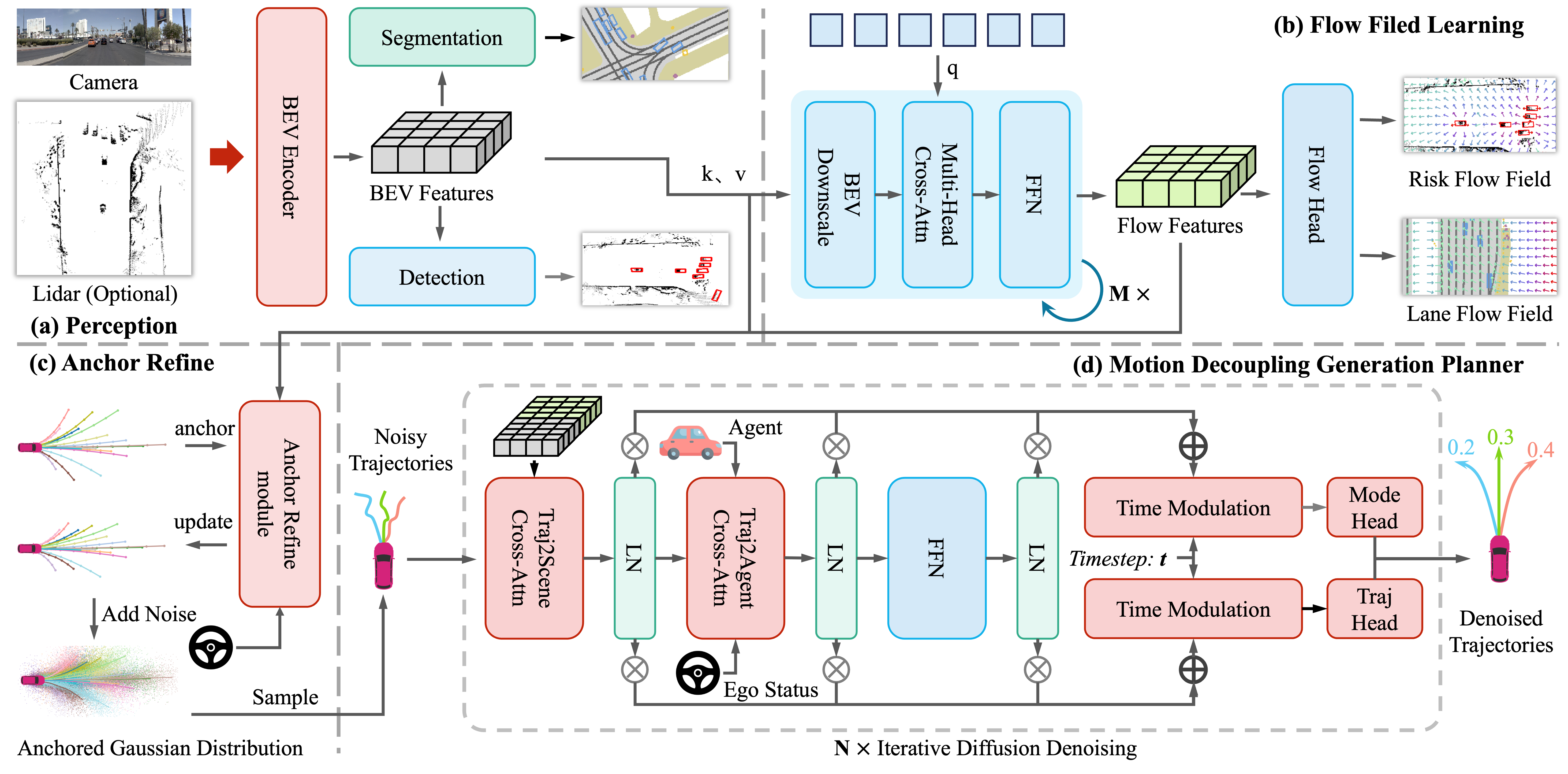}
    \caption{\textbf{The Overall architecture of FlowDrive.} (a) FlowDrive first constructs BEV features from multimodal sensor inputs under the supervision of two auxiliary tasks. (b) Flow Field Learning module leverages BEV features and learnable queries to encode scene-dependent flow-field representations. (c) The learned Flow features are incorporated into an anchor refinement module to adaptively update anchors in a field-aware manner. (d) Finally, the planning module integrates motion decoupling with a diffusion-based generator to produce diverse, multimodal trajectories.}
    \label{fig:framework}
\end{figure*}

In this section, we presents the details of FlowDrive, as illustrated in Fig.~\ref{fig:framework}. FlowDrive begins with a perception module, which extract BEV features from multimodal sensor inputs (Sec.~\ref{sec:preliminary}). These BEV features are then used to learn dense, energy-based flow-field representations that explicitly encode risk and lane priors within the BEV space (Sec.~\ref{sec:flow}). Based on these flow features, an anchor refinement module iteratively adjusts anchor points to align with safe and goal-directed regions (Sec.~\ref{sec:anchor}). Subsequently, FlowDrive integrates feature-level motion decoupling with a conditional diffusion-based generator to generate diverse, feasible trajectory distributions (Sec.~\ref{sec:decoulple}). Finally, we detail the overall training objective (Sec.~\ref{training}).

\subsection{Preliminary}
\label{sec:preliminary}

\textbf{Task Formulation} \quad We consider the problem of end-to-end motion planning for autonomous driving. Given sensor observations (e.g., multi-view camera images or LiDARs), the objective is to generate a safe and feasible future trajectory for the ego vehicle over a planning horizon of \( t_f \) steps, denoted by
\begin{equation}
    \hat{\tau} = \{ \hat{x}_t \}_{t=1}^{t_f}, \quad \hat{x}_t \in \mathbb{R}^3,
    \label{eq:1}
\end{equation}
where each \( \hat{x}_t \) represents the predicted 2D position and heading of the ego vehicle at timestep \( t \). The predicted trajectory should conform to the road topology, avoid static and dynamic obstacles, and reflect plausible high-level driving behaviors (e.g., turning, yielding, or going straight).

\noindent
\textbf{Preception} \quad The perception module serves as the front-end of FlowDrive, transforming raw sensor observations into a compact and semantically rich BEV representation. In this work, we follow the architecture of TransFuser\cite{transdiffuser}, which effectively fuses image and LiDAR modalities through transformer-based attention mechanisms. Specifically, we use front-facing images from three camera views along with LiDAR point clouds, each modality is independently encoded by a separate backbone network. The resulting features are projected into a unified latent space and fused via a multi-stage transformer, enabling hierarchical cross-modal interaction and spatial alignment. The fused representation is then transformed into a top-down BEV feature map \( \mathbf{F}_{\text{BEV}} \in \mathbb{R}^{C \times H \times W} \), which serves as the unified input for subsequent modules. Notably, two auxiliary tasks—semantic map segmentation and object detection—are introduced to supervise the BEV feature learning process, encouraging the network to capture fine-grained static topology and dynamic agent representations.

\begin{figure}[t!]
  \centering
  \begin{subfigure}{0.31\linewidth}
      \centering
      \includegraphics[width=\linewidth]{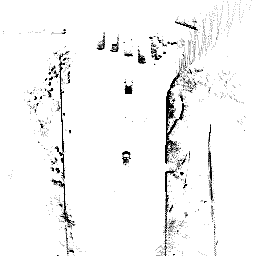}
      \caption{Lidar BEV}
      \label{fig:flow_a}
  \end{subfigure}
  \hfill
  \begin{subfigure}{0.31\linewidth}
      \centering
      \includegraphics[width=\linewidth]{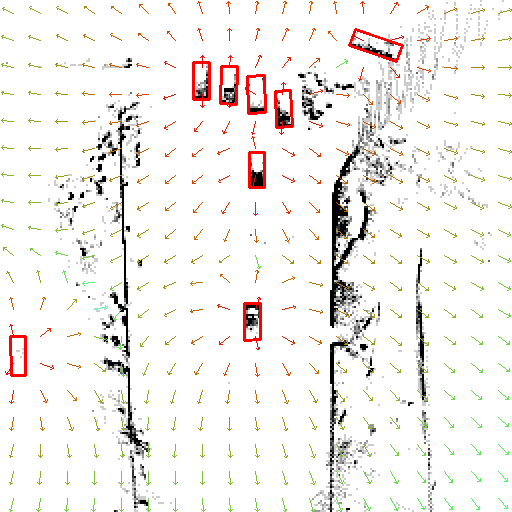}
      \caption{Flow Field}
      \label{fig:flow_b}
  \end{subfigure}
  \hfill
  \begin{subfigure}{0.31\linewidth}
      \centering
      \includegraphics[width=\linewidth]{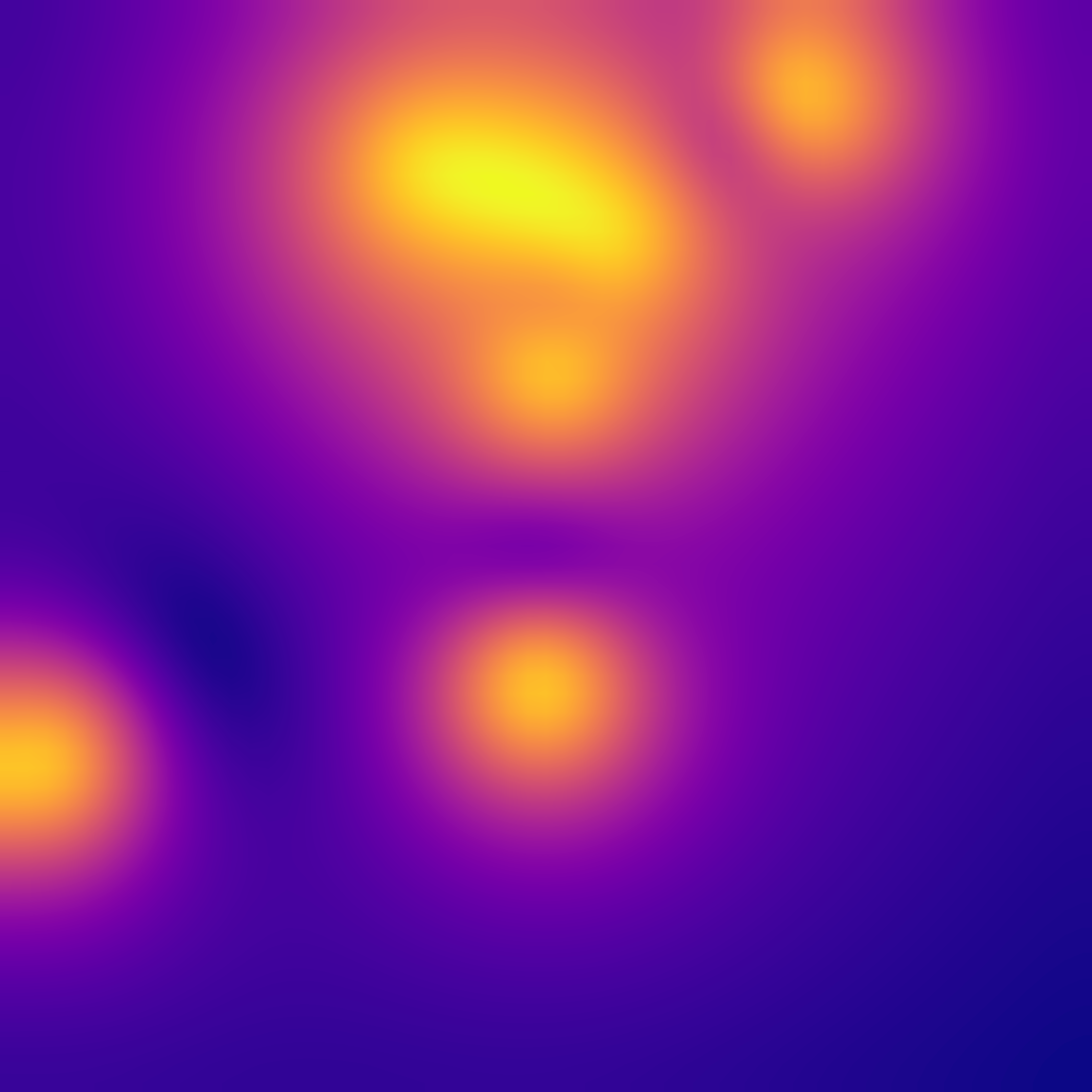}
      \caption{Flow Field Feature}
      \label{fig:flow_c}
  \end{subfigure}
  \caption{\textbf{Visualization of flow field representations in FlowDrive.} \textbf{(a)} BEV projection of LiDAR point clouds showing the static scene geometry. \textbf{(b)} Visualized flow field with risk potential vectors overlaid, guiding the vehicle toward goal-aligned and safe regions \textbf{(c)} Corresponding learned flow features represented as dense energy maps, capturing semantic priors and spatial influence for downstream planning.}
  \label{fig:flow_example}
\end{figure}

\subsection{Flow Filed Learning}
\label{sec:flow}
To inject explicit safety and semantic priors into the planning process, we introduce an energy-based flow field representation embedded in the BEV space (Fig.~\ref{fig:flow_example}). These flow fields encode dense, physically interpretable spatial gradients that provide fine-grained guidance for downstream anchor refinement and trajectory planning. We define two complementary energy fields:
\begin{itemize}
    \item \textbf{Risk Potential Field} \( U_{\text{risk}} \in \mathbb{R}^{H_f \times W_f}\): which assigns higher energy to unsafe regions such as dynamic obstacles, promoting repulsive motion away from them.
    \begin{equation}
        U_{\text{risk}}(u,v) = \sum_{i} \eta \exp\left(-\frac{\| (u,v) - (u_i, v_i) \|^2}{2\sigma^2} \right),
        \label{eq:2}
    \end{equation}
    where \( (u_i, v_i) \) is the position of agent \( i \), \( \eta \) is a risk weight, and \( \sigma \) controls the spatial spread.
    \item \textbf{Lane Attraction Field} \( U_{\text{lane}} \in \mathbb{R}^{H_f \times W_f} \): which assigns lower energy to drivable areas and goal-directed regions, encouraging attractive motion toward safe and feasible paths.
    \begin{equation}
        U_{\text{lane}}(u, v) = \frac{1}{2} k_{\text{lat}} d(u,v)^2 + k_{\text{lon}} \left( L - s(u,v) \right),
        \label{eq:3}
    \end{equation}
    where \( d(u,v) \) represents lateral distance from point \( (u,v) \) to the nearest lane centerline, \( s(u,v) \) is longitudinal arc length from the start of the lane to the projection of \( (u,v) \), \( L \) is total arc length of the centerlane, \( k_{\text{lat}}, k_{\text{lon}} \) represent lateral and longitudinal weighting coefficients. Together, these fields construct a continuous vector landscape that encodes safety and driving intent in the BEV domain.
\end{itemize}

To model these fields, we first downsample the BEV features \( \mathbf{F}_{\text{BEV}} \in \mathbb{R}^{C \times H \times W} \) to obtain a compact spatial representation. Then we introduce a set of learnable queries \( Q \in \mathbb{R}^{N \times D} \), which interact with the downsampled features through multi-head attention and feed-forward networks to extract flow features \( \mathbf{F}_{\text{flow}} \in \mathbb{R}^{C \times H \times W} \). These features are then decoded into the predicted energy maps:
\begin{equation}
    \hat{U}_{\text{risk}},\ \hat{U}_{\text{lane}} = \text{MLP}(\mathbf{F}_{\text{flow}}),
    \label{eq:4}
\end{equation}
where  \( \hat{U}_{\text{risk}} \), \( \hat{U}_{\text{lane}} \) represent predicted risk potential field and lane attraction field.

\subsection{Flow-Aware Anchor Refinement}
\label{sec:anchor}

\begin{figure}[t]
    \centering
    \includegraphics[width=0.8\linewidth]{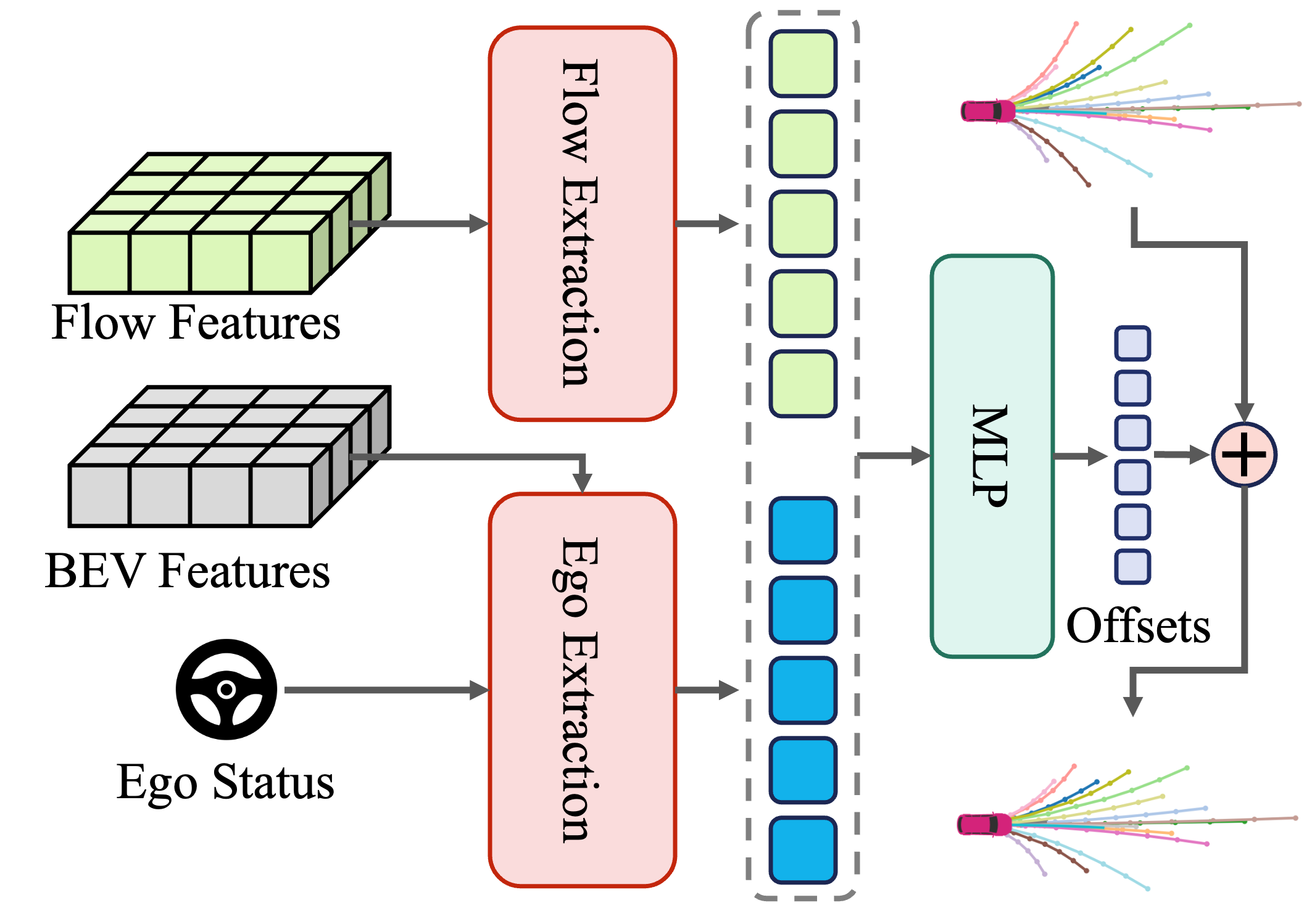}
    \caption{Flow-Aware Anchor Refine Module}
    \label{fig:anchor_refine}
\end{figure}

To improve the spatial alignment between planned trajectories and the underlying driving context, we propose a flow-aware anchor refinement module that adjusts the initial anchor points in a scene-adaptive manner. We begin with a set of pre-defined anchor trajectories \(\mathcal{A} = \{\tau_i\}_{i=1}^N\) clustered by K-Means, each initialized from a fixed spatial position and orientation. However, due to the diversity of road layouts and dynamic scenes, fixed anchors often fail to align with safe or goal-directed regions. Initial anchors are represented as \( \mathcal{A} \in \mathbb{R}^{N \times t_f \times 2} \), where \( N \) denotes the number of anchors, \( t_f \) is the feature dimension. The anchors are updated as follows (Fig.~\ref{fig:anchor_refine}):

\begin{equation}
    \tilde{\mathcal{A}_0} = \text{MLP}(E_{flow}(\textbf{F}_{flow}), E_{ego}(\mathbf{F}_{\text{ego}}, \mathbf{F}_{\text{flow}})) + \mathcal{A},
    \label{eq:5}
\end{equation}
where $E_{flow}$ denotes a global average pooling operation followed by flattening, $E_{ego}$ represents a transformer-based encoder, and $\mathbf{F}_{\text{ego}}$ corresponds to the ego status.

The refined anchor embeddings capture localized semantic and safety-aware context from the flow field. These refined representations are used to condition the downstream trajectory generation planner. This mechanism allows the anchor space to adapt dynamically to the driving scene without requiring explicit coordinate updates, and provides a fully differentiable refinement pathway compatible with the end-to-end training pipeline.

\subsection{Motion Decoupling Generation Planner}
\label{sec:decoulple}
Inspired by DiffusionDrive \cite{diffusiondrive} and DiffusionPlanner \cite{diffusionplanner}, we adopt a conditional diffusion framework to model the multi-modal trajectory distribution. Differently, to mitigate gradient interference and representation entanglement between mode prediction and trajectory generation, we introduce a feature-level decoupling strategy that separates their learning objectives and enables targeted feature extraction for each task.

Following the standard formulation of Denoising Diffusion Probabilistic Models (DDPM) \cite{ddpm}, we first define a forward process that gradually adds Gaussian noise to the refined anchors \(\tilde{\mathcal{A}_0}\). At each timestep \( t \in \{1,...,T\}\), the noisy anchors \(\tilde{\mathcal{A}_t}\) is computed as:
\begin{equation}
    \tilde{\mathcal{A}_t} = \sqrt{\bar{\alpha}_t} \cdot \tilde{\mathcal{A}_0} + \sqrt{1 - \bar{\alpha}_t} \cdot \boldsymbol{\epsilon}, \quad \boldsymbol{\epsilon} \sim \mathcal{N}(0, \mathbf{I}),
    \label{eq:6}
\end{equation}
where \(\bar{\alpha}_t = \prod_{s=1}^{t} \alpha_s\) is the cumulative product of predefined variance schedule coefficients, and \(\boldsymbol{\epsilon}\) is sampled from the standard Gaussian distribution. 

Noisy anchors \( \tilde{\mathcal{A}_t} \) are treated as trajectory queries, which attend to contextual features through a cross-attention mechanism, followed by projection into a shared latent space via a feed-forward network. To further disentangle the learning of motion mode and trajectory generation, we introduce two sets of learnable gating queries, \( g_1 \in \mathbb{R}^{N \times 3} \) and \( g_2 \in \mathbb{R}^{N \times 3} \), which selectively aggregate information from intermediate representations, enabling task-specific feature routing. Finally, task-specific heads are employed for mode prediction and trajectory generation. At inference time,, similarly to DiffusionDrive \cite{diffusiondrive}, we adopt the DDIM sampling procedure \cite{ddim} to iteratively refine the trajectory through a deterministic denoising process.

\begin{equation}
    \tilde{\mathcal{A}}_{t-1} = \sqrt{\bar{\alpha}_{t-1}} \cdot \hat{\mathcal{A}}_0 + \sqrt{1 - \bar{\alpha}_{t-1}} \cdot 
    \left( 
        \frac{\tilde{\mathcal{A}}_t - \sqrt{\bar{\alpha}_t} \cdot \hat{\mathcal{A}}_0}{\sqrt{1 - \bar{\alpha}_t}} 
    \right),
    \label{eq:7}
\end{equation}

\begin{equation}
    \{ \hat{s}_k, \hat{\tau}_k \}_{k=1}^{N} = \tau_\theta(\tilde{\mathcal{A}}_0, z),
    \label{eq:8}
\end{equation}
where \( \tilde{\mathcal{A}}_t \in  \mathbb{R}^{N \times t_f \times 2} \)denotes the noisy anchors at timestep \( t \), and \( \tilde{\mathcal{A}}_0 \)
is the denoised anchors by the model. \( \bar{\alpha}_t \in (0,1) \) represents the cumulative noise schedule. \( \tau_\theta(\cdot) \) denotes the motion decoulping generation planner that outputs \( N\) trajectory candidates \( \{ \hat{\tau}_k \}_{k=1}^{N} \) along with their confidence scores \( \{ \hat{s}_k \}_{k=1}^{N} \).

\begin{table*}[!ht]
  \centering
  \setlength{\tabcolsep}{0.13cm} 
  \begin{tabular}{l|c|c|ccccccccc|c}
    \toprule
    \textbf{Method} & \textbf{Backbone} & \textbf{Sensor} & \textbf{NC $\uparrow$} & \textbf{DAC $\uparrow$} & \textbf{DDC $\uparrow$} & \textbf{TL $\uparrow$} & \textbf{EP $\uparrow$} & \textbf{TTC $\uparrow$} & \textbf{LK $\uparrow$} & \textbf{HC $\uparrow$} & \textbf{EC $\uparrow$} & \textbf{EPDMS $\uparrow$}\\
    \midrule
    Human Agent &- &Img. &100 &100 &99.8 &100 &87.4 &100 &100 &98.1 &90.1 &90.3 \\
    Ego Status MLP &- &Img. &93.1 &77.9 &92.7 &99.6 &86.0 &91.5 &89.4 &98.3 &85.4 &64.0\\
    \midrule
    Transfuser\cite{transfuser} &ResNet34 &Img. &96.9 &89.9 &97.8 &99.7 &87.1 &95.4 &92.7 &\textbf{98.3} &87.2 &76.7\\
    VADv2\cite{vadv2} &ResNet34 &Img. &97.3 &91.7 &77.6 &92.7 &\textbf{100} &\textbf{99.9} &\textbf{98.2} &66.0 &\textbf{97.4} &76.6\\
    HydraMDP\cite{hydramdp} &ResNet34 &Img. &97.5 &96.3 &80.1 &93.0 &\textbf{100} &\textbf{99.9} &\textbf{98.3} &65.5 &\textbf{97.4} &79.8\\
    HydraMDP++\cite{hydramdp++} &ResNet34 &Img. &97.2 &\textbf{97.5} &\underline{99.4} &99.6 &83.1 &96.5 &94.4 &98.2 &70.9 &81.4\\
    Diffusiondrive\cite{diffusiondrive} &ResNet34 &Img. &\underline{98.0} &96.0 &\textbf{99.5} &\textbf{99.8} &87.7 &97.1 &97.2 &\textbf{98.3} &\underline{87.6} &84.3\\
    DriveSuprim\cite{drivesuprim} &ResNet34 &Img. &97.5 &\underline{96.5} &\underline{99.4} &99.6 &\underline{88.4} &96.6 &95.5 &\textbf{98.3} &77.0 &83.1\\    
    PRIX\cite{prix} &ResNet34 &Img. &\underline{98.0} &95.6 &\textbf{99.5} &\textbf{99.8} &87.4 &97.2 &97.1 &\textbf{98.3} &\underline{87.6} &84.2\\
    \rowcolor{blue!10}\textbf{FlowDrive(our)} &ResNet34 &Img. &\textbf{98.3} &96.2 &\textbf{99.5} &\textbf{99.8} &87.7 &\underline{97.4} &\underline{97.4} &\textbf{98.3} &\underline{87.6} &\textbf{84.9}\\
    \midrule
    Diffusiondrive\cite{diffusiondrive} &V2-99 &Img. &98.2 &96.3 &\textbf{99.6} &\underline{99.8} &87.5 &\underline{97.5} &\underline{97.1} &\textbf{98.3} &\textbf{87.7} &85.0\\
    HydraMDP++\cite{hydramdp++} &V2-99 &Img. &\underline{98.4} &\textbf{98.0} &99.4 &\underline{99.8} &87.5 &97.7 &95.3 &\textbf{98.3} &77.4 &85.1\\
    DriveSuprim\cite{drivesuprim} &V2-99 &Img. &97.8 &\underline{97.9} &\underline{99.5} &\textbf{99.9} &\textbf{90.6} &97.1 &96.6 &\textbf{98.3} &77.9 &86.0\\
    \rowcolor{blue!10} \textbf{FlowDrive(our)} &V2-99 &Img. &\textbf{98.5} &97.4 &\textbf{99.6} &\textbf{99.9} &\underline{87.7} &\textbf{97.9} &\textbf{97.8} &\textbf{98.3} &\underline{87.6} &\textbf{86.3}\\
    \bottomrule
  \end{tabular}
  \caption{Comparison on NAVSIM v2 navtset split with EPDMS metrics. Results are grouped by backbone types and sensor inputs. The \textbf{best} and the \underline{second best} results are denoted by \textbf{bold} and \underline{underline}.}
  \label{tab:quantitive}
  \vspace{-10pt}
\end{table*}

\subsection{Training Objective}
\label{training}
To optimize FlowDrive, we define a composite loss function that supervises the model from four complementary perspectives: trajectory planning, anchor refinement, flow field modeling, and auxiliary perception tasks. The overall training objective is:
\begin{equation}
    \mathcal{L} = \mathcal{L}_{\text{plan}} + \mathcal{L}_{\text{anchor}} + \mathcal{L}_{\text{flow}} + \mathcal{L}_{\text{aux}},
    \label{eq:9}
\end{equation}
The planning loss includes both mode classification and trajectory generation. Specifically, we define:
\begin{equation}
    \mathcal{L}_{\text{plan}} = \sum_{k=1}^{N} \left[ y_k \cdot \lambda_\text{traj} \cdot \mathcal{L}_1(\hat{\tau}_k, \tau_{\text{gt}}) + \lambda_\text{mode} \cdot \mathcal{L}_{BCE}(\hat{s}_k, y_k) \right],
\end{equation}
where \( y_k = 1 \) if the \(k\)-th predicted trajectory \( \hat{\tau}_k \) is the closest to the ground-truth trajectory \( \tau_{\text{gt}} \), and \( y_k = 0 \) otherwise.
Anchor regression loss is defined to ensure the refined anchors remain spatially consistent with the ground-truth trajectories:
\begin{equation}
    \mathcal{L}_{\text{anchor}} = \frac{1}{N} \sum_{i=1}^{N} y_k \cdot \lambda_\text{anchor} \cdot \mathcal{L}_1(\tilde{\mathcal{A}}_k, \tau_{\text{gt}}),
\end{equation}
Flow loss use MSE function to guide the learning of interpretable flow field representations.
\begin{equation}
    \mathcal{L}_{\text{flow}} = \lambda_\text{traj} \cdot ( \mathcal{L}_{mse}(\hat{U}_{\text{risk}}, U_{\text{risk}}) + \mathcal{L}_{mse}(\hat{U}_{\text{lane}}, U_{\text{lane}}) ),
\end{equation}
Auxiliary perception losses are employed to enhance scene understanding, we include two auxiliary tasks: semantic segmentation and object detection. The loss is defined as:
\begin{equation}
    \mathcal{L}_{\text{aux}} = \lambda_\text{seg} \cdot \mathcal{L}_{\text{seg}} + \lambda_\text{det} \cdot \mathcal{L}_{\text{det}}.
\end{equation}
where \( \mathcal{L}_{\text{seg}} \) is a pixel-wise cross-entropy loss for BEV semantic segmentation, and \( \mathcal{L}_{\text{det}} \) is a standard detection loss (classification + box regression). 
All the coefficients \(\lambda_*\) are empirically selected as hyperparameters o balance the contributions of different loss components during training.

\section{Experiments}
\label{sec:experiment}

\subsection{Dataset and Metrics}
\label{sec:dataset}
We conduct all experiments on the NAVSIM \cite{navsim} dataset, a large-scale simulation benchmark designed for evaluating end-to-end autonomous driving systems in complex urban environments. NAVSIM provides diverse traffic scenarios with multi-agent interactions, traffic signals, and structured road geometry. Each sample includes multi-view camera images, optional LiDAR data, HD maps, and ground-truth trajectories for both ego and surrounding agents. To quantitatively evaluate planning performance, we adopt The Extended Predictive Driver Model Score (EPDMS) from NAVSIM v2, which offers a comprehensive set of metrics for assessing trajectory quality, safety, and driving rationality. Specifically, EPDMS evaluates generated trajectories across multiple aspects, including No at-fault Collisions (NC), Drivable Area Compliance (DAC), Driving Direction Compliance (DDC), Traffic Light Compliance (TLC), Ego Progress (EP), Time to Collision (TTC), Lane Keeping (LK), History Comfort (HC), Extended Comfort (EC).

\subsection{Implementation Details}

We experiment with different image backbones, including ResNet-34 \cite{resnet} and VoVNet \cite{vovnet}, where input images are resized to a resolution of \(2048 \times 512\). The perception architecture follows TransFuser \cite{transfuser}, which fuses multi-view camera features and optional LiDAR inputs into BEV representations through a multi-stage transformer. The Flow Field Learning module consists of a single downscale layer, followed by cross-attention and an FFN to extract flow features. A 1-layer MLP is applied to predict the flow field maps. For the Anchor Refinement module, we employ a 1-layer transformer that adaptively updates anchors, which are initialized from a set of 20 clustered trajectories. In Motion Decoupled Generation Planner, two cross-attention layers are employed to aggregate scene and agent features, followed by a FFN to project into the latent space. To disentangle motion intent and trajectory denoising, we introduce two sets of learnable gating queries that selectively integrate intermediate features and apply temporal modulation. The resulting features are passed into two task-specific heads, each implemented as a 2-layer MLP, for mode prediction and trajectory generation, respectively. FlowDrive is trained on 8 NVIDIA H20 GPUs using the Adam optimizer for 100 epochs, with a batch size of 8 and an initial learning rate of \(1.5 \times 10^{-4}\).

\begin{table*}
  \centering
  \setlength{\tabcolsep}{0.22cm} 
  \begin{tabular}{ccc|ccccccccc|c}
    \toprule
    \textbf{Flow.} & \textbf{Adap.} & \textbf{Motion.} & \textbf{NC $\uparrow$} & \textbf{DAC $\uparrow$} & \textbf{DDC $\uparrow$} & \textbf{TL $\uparrow$} & \textbf{EP $\uparrow$} & \textbf{TTC $\uparrow$} & \textbf{LK $\uparrow$} & \textbf{HC $\uparrow$} & \textbf{EC $\uparrow$} & \textbf{EPDMS $\uparrow$}\\
    \midrule
    \rowcolor{blue!10} \ding{51} &\ding{51} &\ding{51} &98.5 &97.4 &99.6 &99.9 &87.7 &97.9 &97.8 &98.3 &87.6 &86.3 \\
    \midrule
    \ding{55} &\ding{51} &\ding{51} &98.5 &97.1 &99.5 &99.9 &87.4 &97.9 &97.6 &98.3 &87.0 &85.8(\textcolor{Red}{-0.5}) \\
    \ding{51} &\ding{55} &\ding{51} &98.4 &97.2 &99.6 &99.8 &87.5 &97.9 &97.7 &98.3 &86.8 &85.9(\textcolor{Red}{-0.4}) \\
    \ding{51} &\ding{51} &\ding{55} &98.5 &97.2 &99.6 &99.8 &87.6 &97.9 &97.7 &98.3 &87.7 &86.1(\textcolor{Red}{-0.2}) \\
    \bottomrule
  \end{tabular}
  \caption{Ablation study on the effectiveness of the proposed modules: \textbf{Flow.} (Flow Field Learning), \textbf{Adap.} (Anchor Refine module), and \textbf{Motion.} (Motion Decoupling).}
  \label{tab:module}
\end{table*}

\begin{table*}
  \centering
  \setlength{\tabcolsep}{0.22cm} 
  \begin{tabular}{cccc|ccccccccc|c}
    \toprule
    \textbf{$\sigma$} & \textbf{$\eta$} & \textbf{$k_{lat}$} & \textbf{$k_{lon}$} & \textbf{NC $\uparrow$} & \textbf{DAC $\uparrow$} & \textbf{DDC $\uparrow$} & \textbf{TL $\uparrow$} & \textbf{EP $\uparrow$} & \textbf{TTC $\uparrow$} & \textbf{LK $\uparrow$} & \textbf{HC $\uparrow$} & \textbf{EC $\uparrow$} & \textbf{EPDMS $\uparrow$}\\
    \midrule
    40.0 &1.0 &1.0 &10.0 &98.5 &97.2 &99.6 &99.8 &87.6 &97.8 &97.7 &98.3 &87.9 &\cellcolor{blue!10}86.1\\
    20.0 &1.0 &1.0 &10.0 &98.5 &97.1 &99.6 &99.8 &87.7 &97.8 &97.5 &98.3 &87.7 &\cellcolor{blue!10}85.9\\
    10.0 &1.0 &1.0 &10.0 &98.5 &97.4 &99.6 &99.9 &87.7 &97.9 &97.8 &98.3 &87.6 &\cellcolor{blue!10}86.3\\
    10.0 &0.5 &1.0 &10.0 &98.3 &97.5 &99.6 &99.8 &87.7 &97.7 &97.8 &98.3 &87.5 &\cellcolor{blue!10}86.2\\
    \bottomrule
  \end{tabular}
  \caption{Results with different parameter configurations of Flow Field. \(\eta\) is a risk weight, \(\sigma\) controls the spatial influence, \( k_{\text{lat}}, k_{\text{lon}} \) represent lateral and longitudinal weighting coefficients in Lane Attraction Field.}
  \label{tab:parameter}
\end{table*}


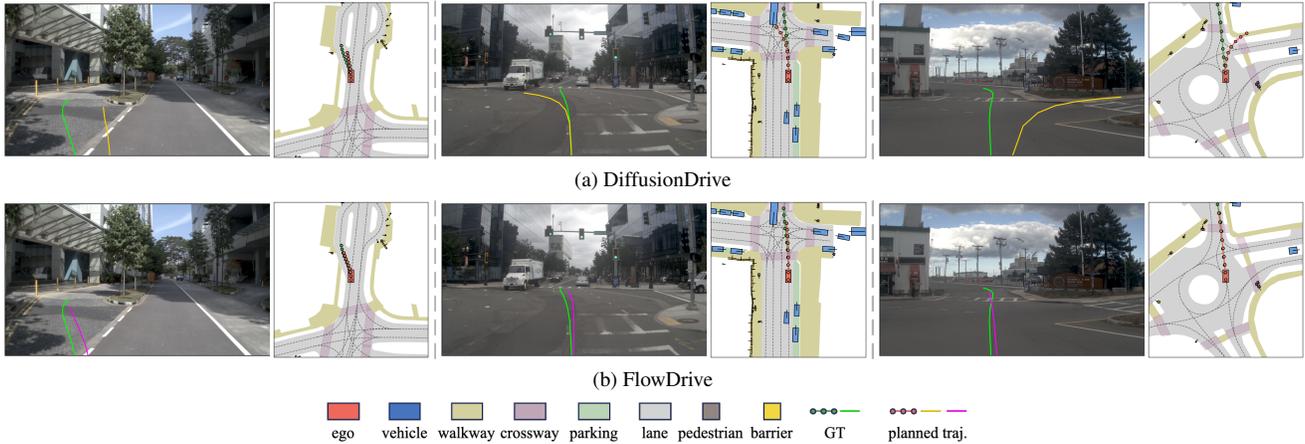
\begin{figure*}[h!]
  \centering
  \begin{subfigure}{\linewidth}
      \centering
      \includegraphics[height=0.123\linewidth]{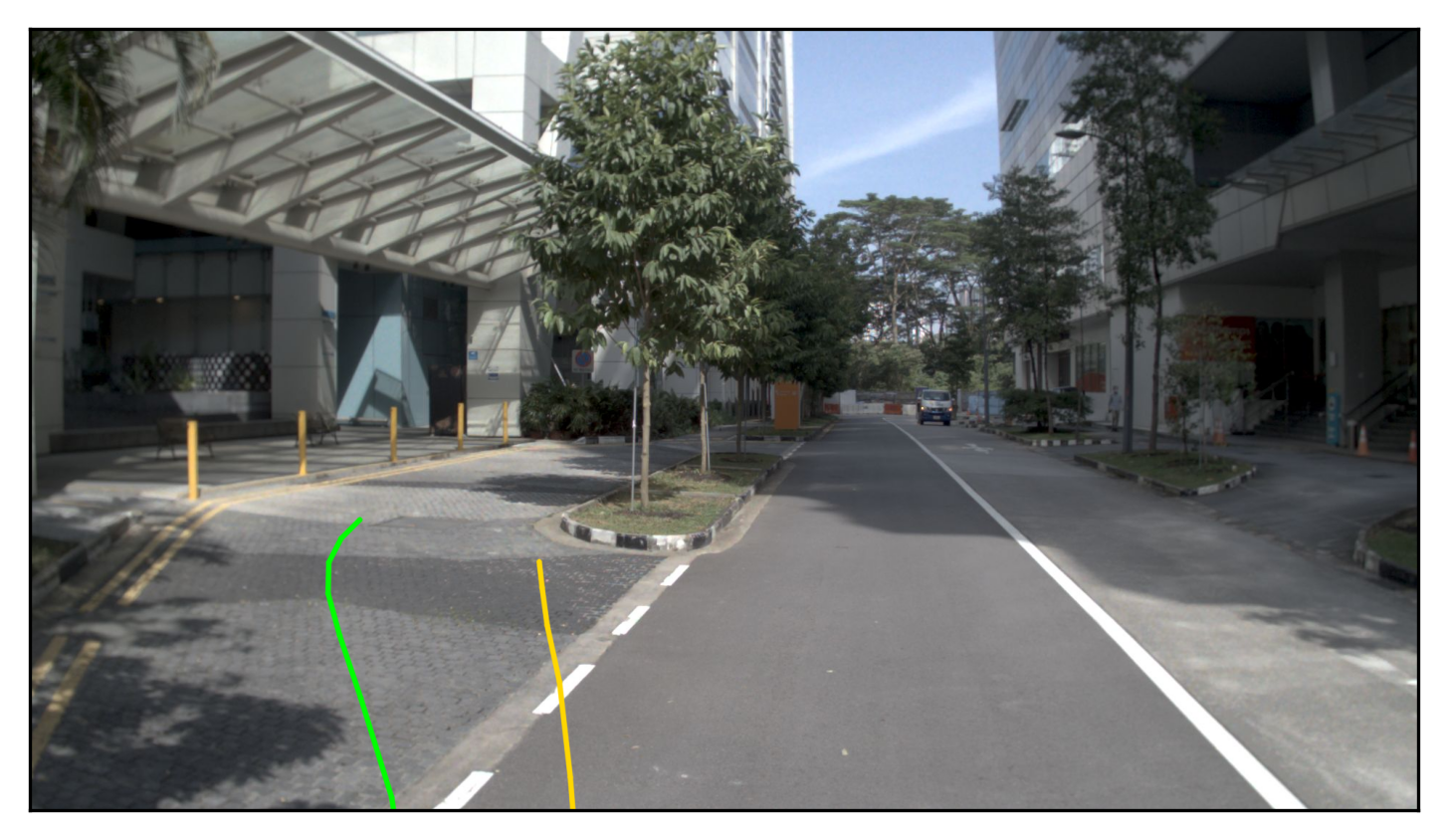}
      \hspace{-8pt}
      \includegraphics[height=0.123\linewidth]{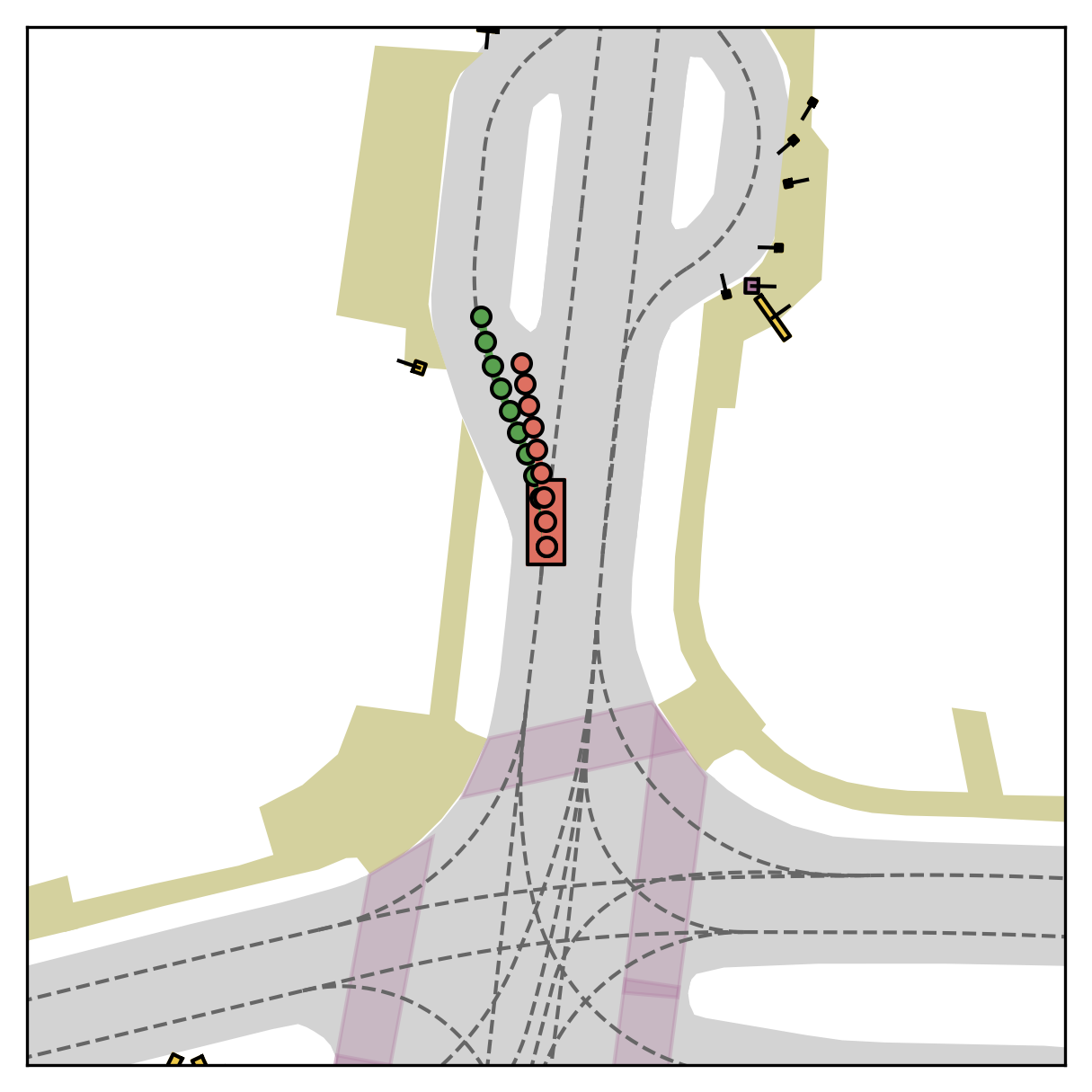}
      \hspace{-4pt}
        \begin{tikzpicture}
            \draw[dash pattern=on 6pt off 3pt, line width=1pt, color=black!30] (0,0) -- (0,0.123\linewidth);
        \end{tikzpicture}
      \hspace{-5pt}
      \includegraphics[height=0.123\linewidth]{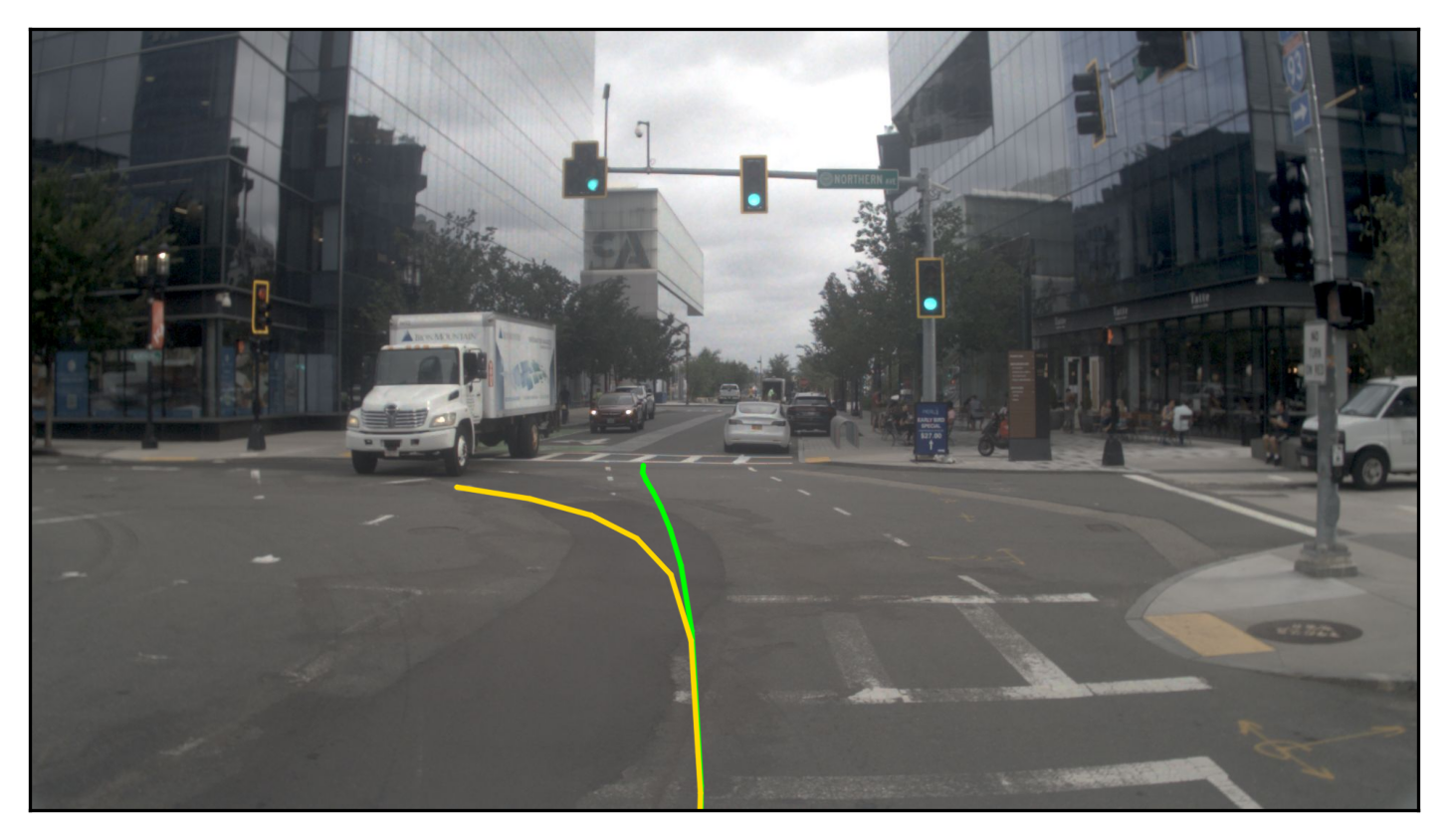}
      \hspace{-8pt}
      \includegraphics[height=0.123\linewidth]{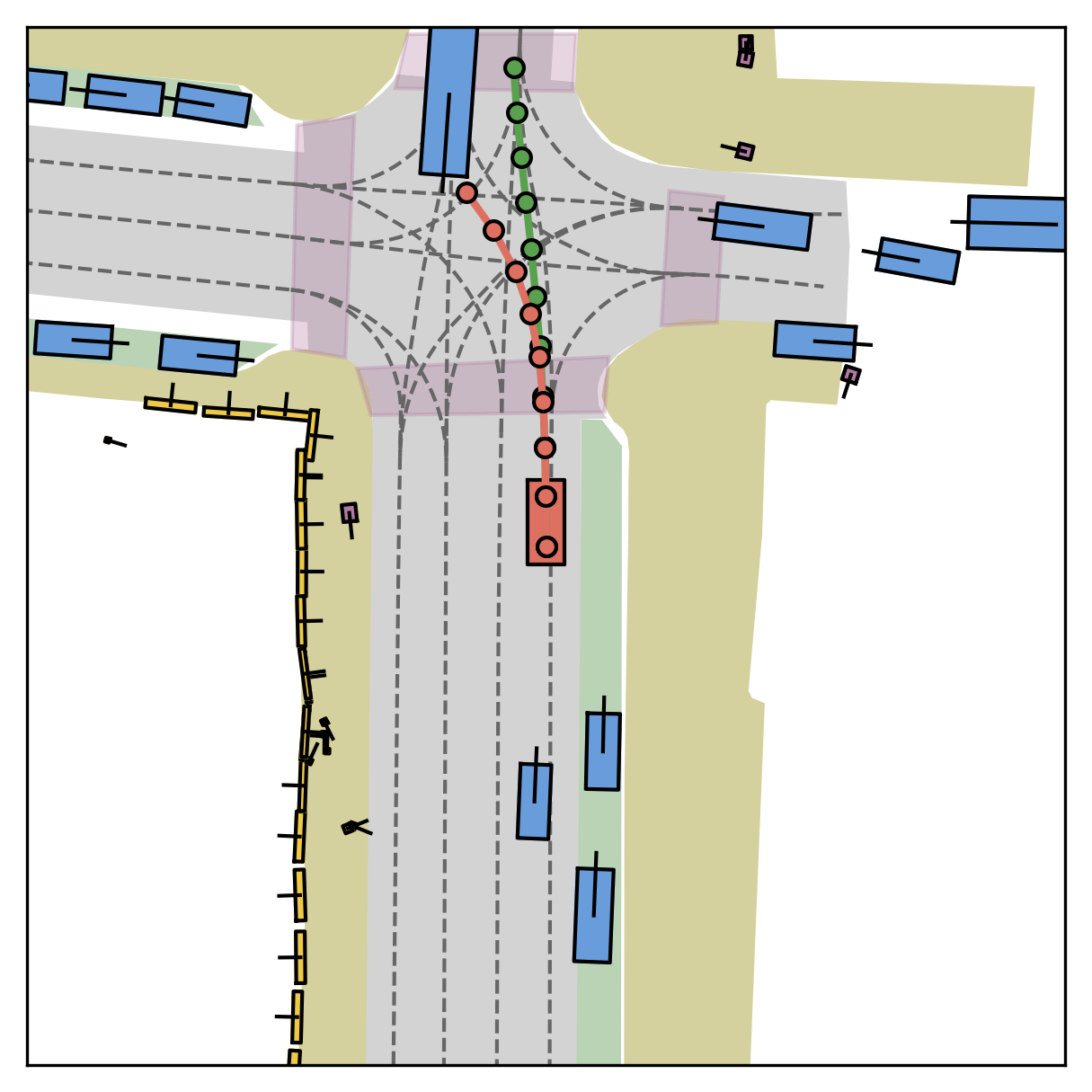}
      \hspace{-4pt}
        \begin{tikzpicture}
            \draw[dash pattern=on 6pt off 3pt, line width=1pt, color=black!30] (0,0) -- (0,0.123\linewidth);
        \end{tikzpicture}
      \hspace{-5pt}
      \includegraphics[height=0.123\linewidth]{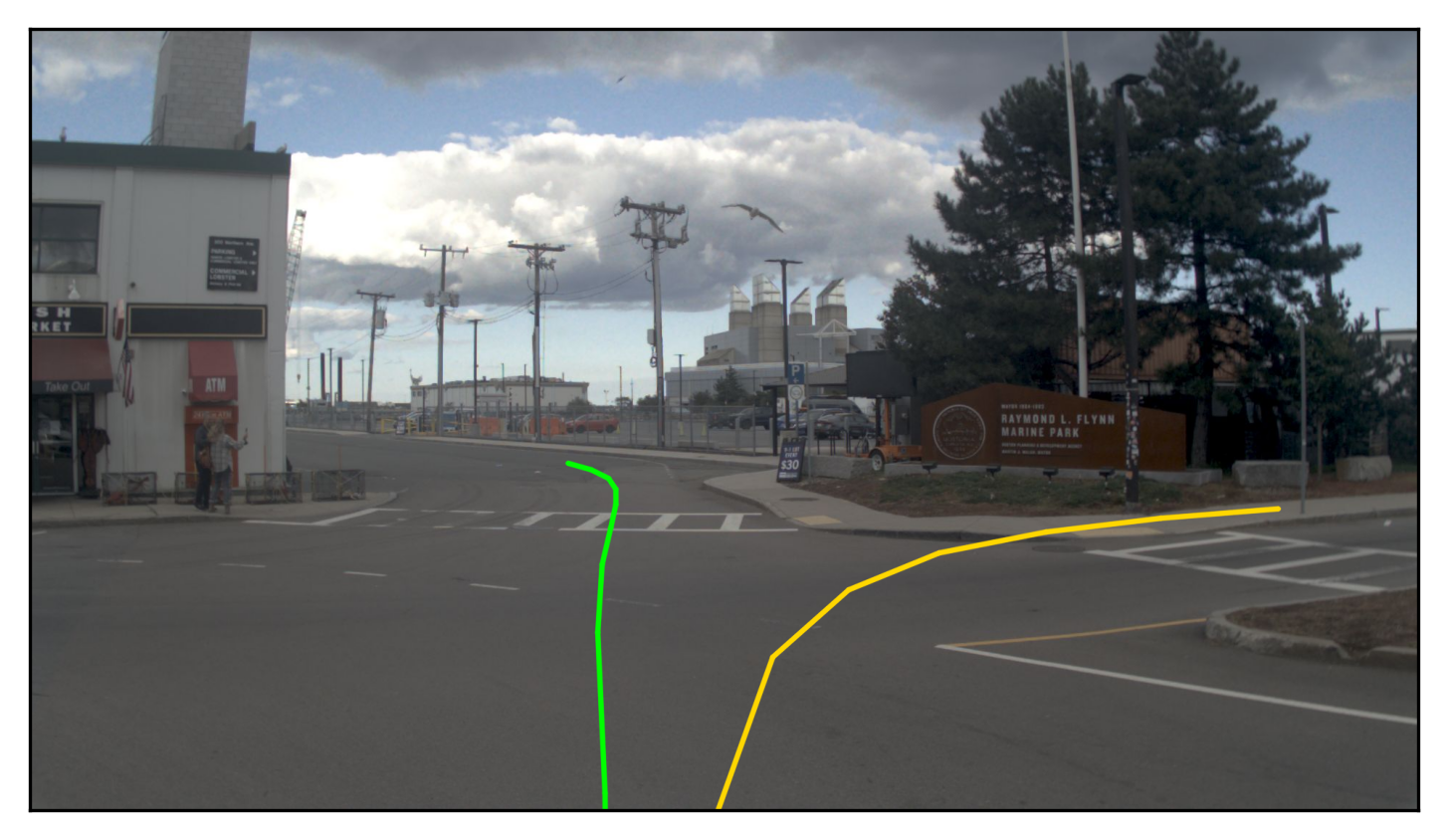}
      \hspace{-8pt}
      \includegraphics[height=0.123\linewidth]{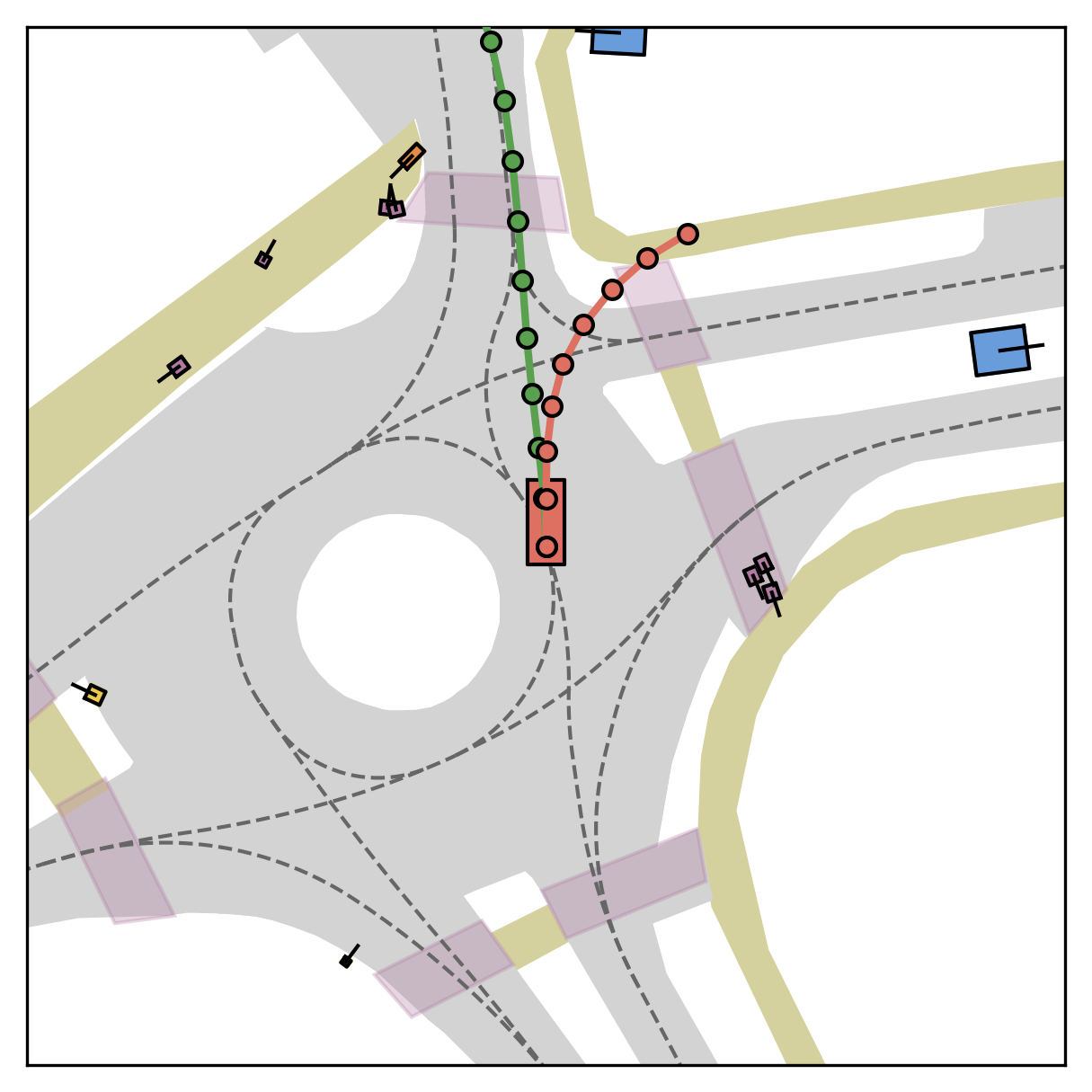}
      \caption{DiffusionDrive}
      \label{fig:v_a}
  \end{subfigure}
    \begin{subfigure}{\linewidth}
      \centering
      \includegraphics[height=0.123\linewidth]{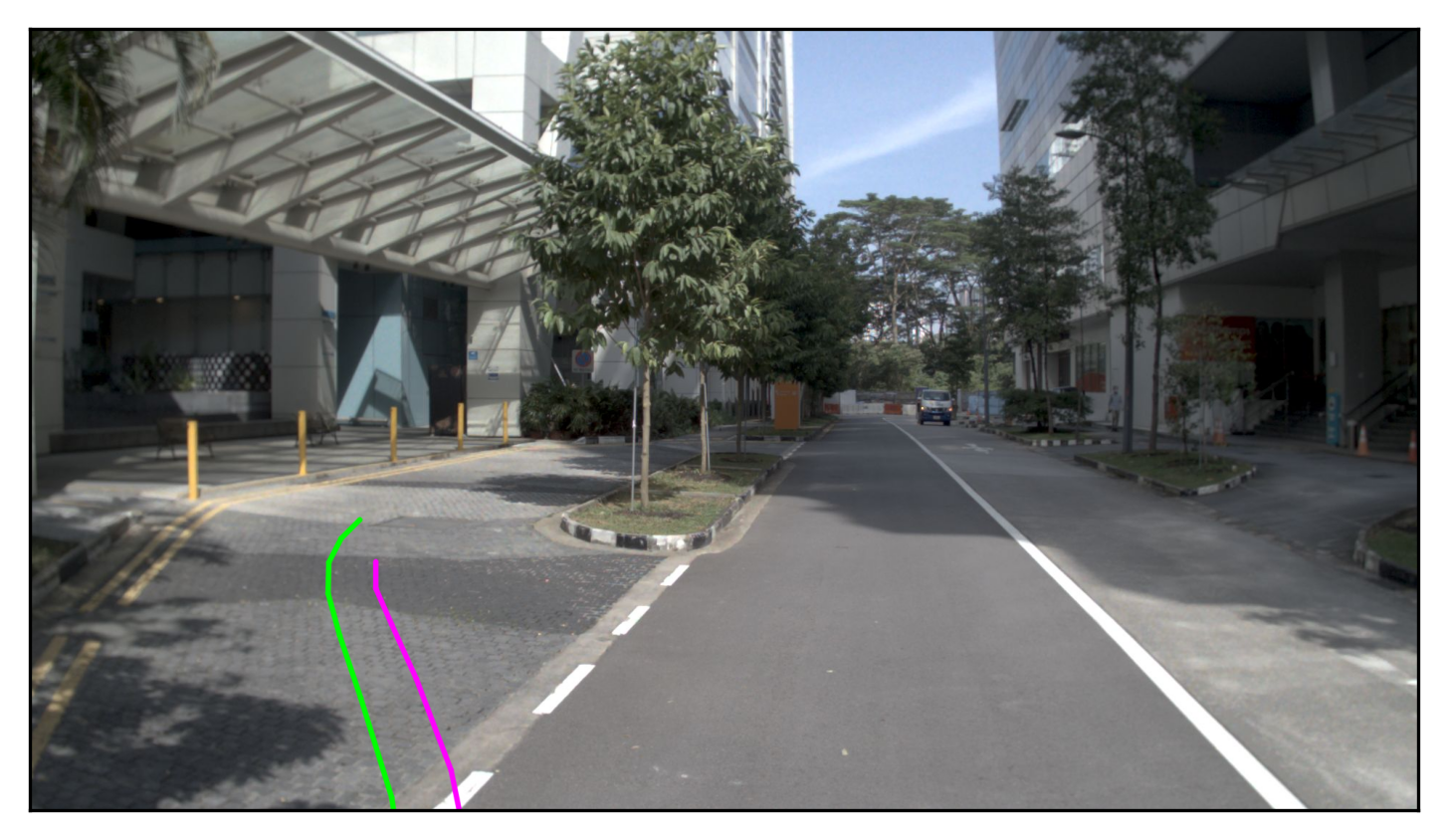}
      \hspace{-8pt}
      \includegraphics[height=0.123\linewidth]{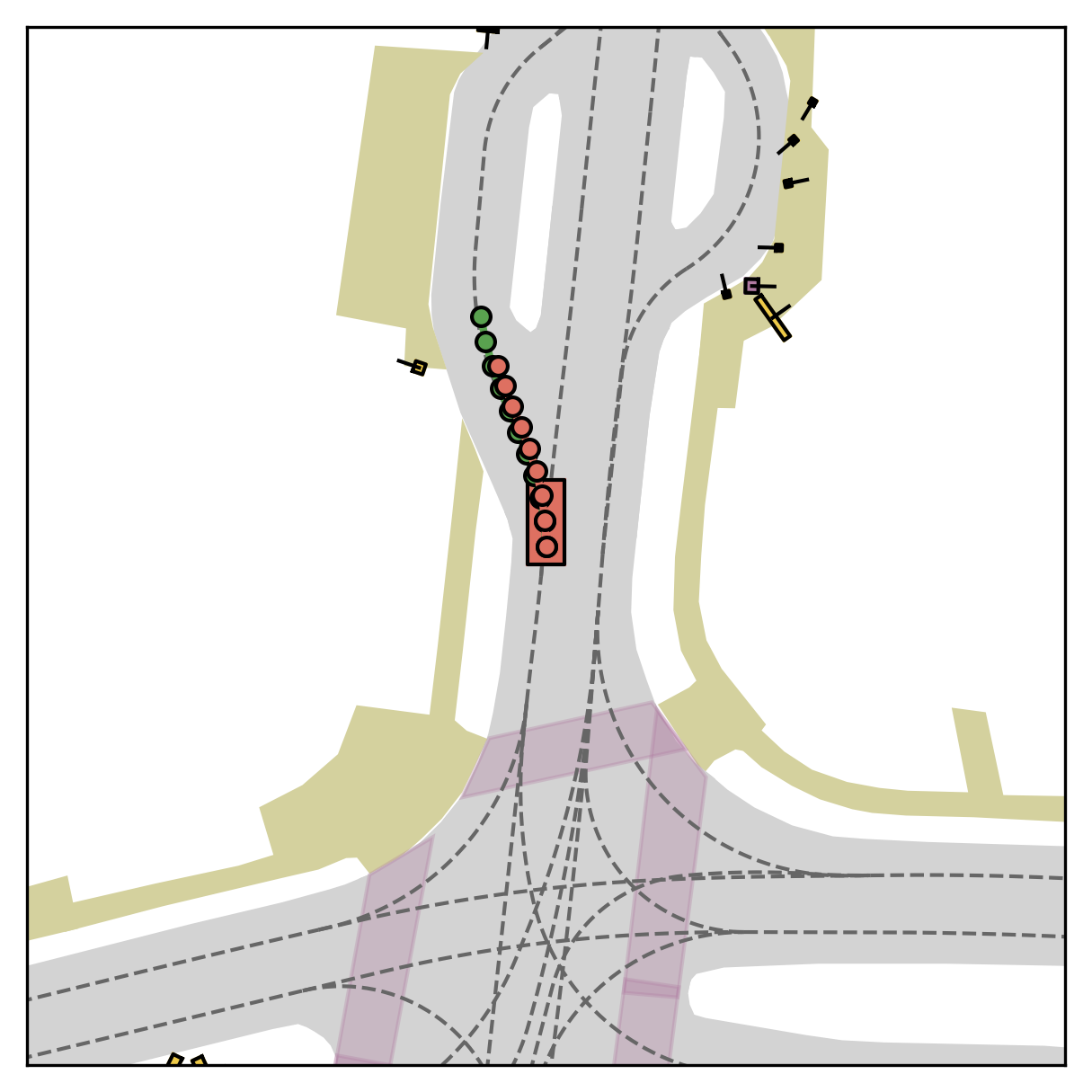}
      \hspace{-4pt}
        \begin{tikzpicture}
            \draw[dash pattern=on 6pt off 3pt, line width=1pt, color=black!30] (0,0) -- (0,0.123\linewidth);
        \end{tikzpicture}
      \hspace{-5pt}
      \includegraphics[height=0.123\linewidth]{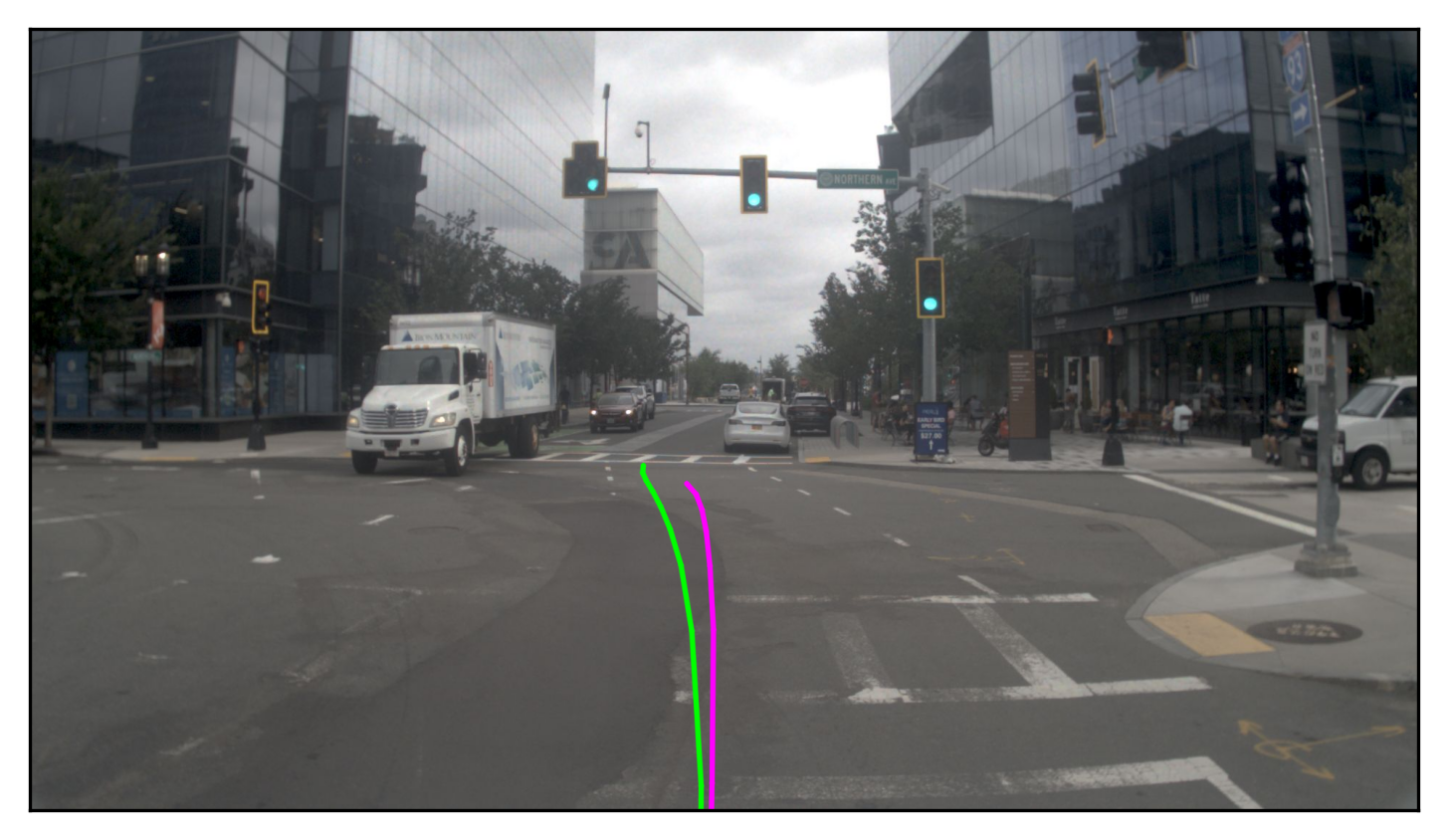}
      \hspace{-8pt}
      \includegraphics[height=0.123\linewidth]{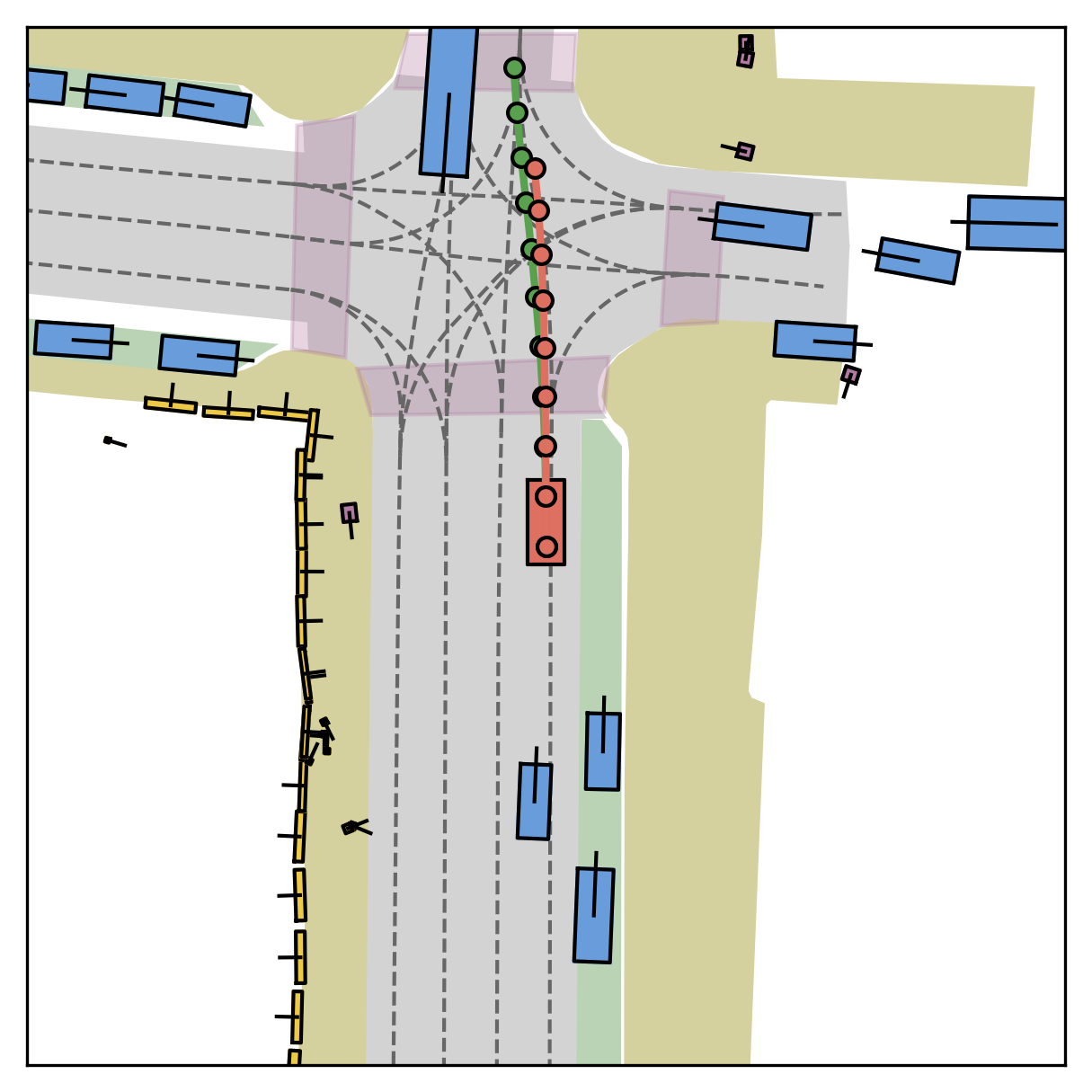}
      \hspace{-4pt}
        \begin{tikzpicture}
            \draw[dash pattern=on 6pt off 3pt, line width=1pt, color=black!30] (0,0) -- (0,0.123\linewidth);
        \end{tikzpicture}
      \hspace{-5pt}
      \includegraphics[height=0.123\linewidth]{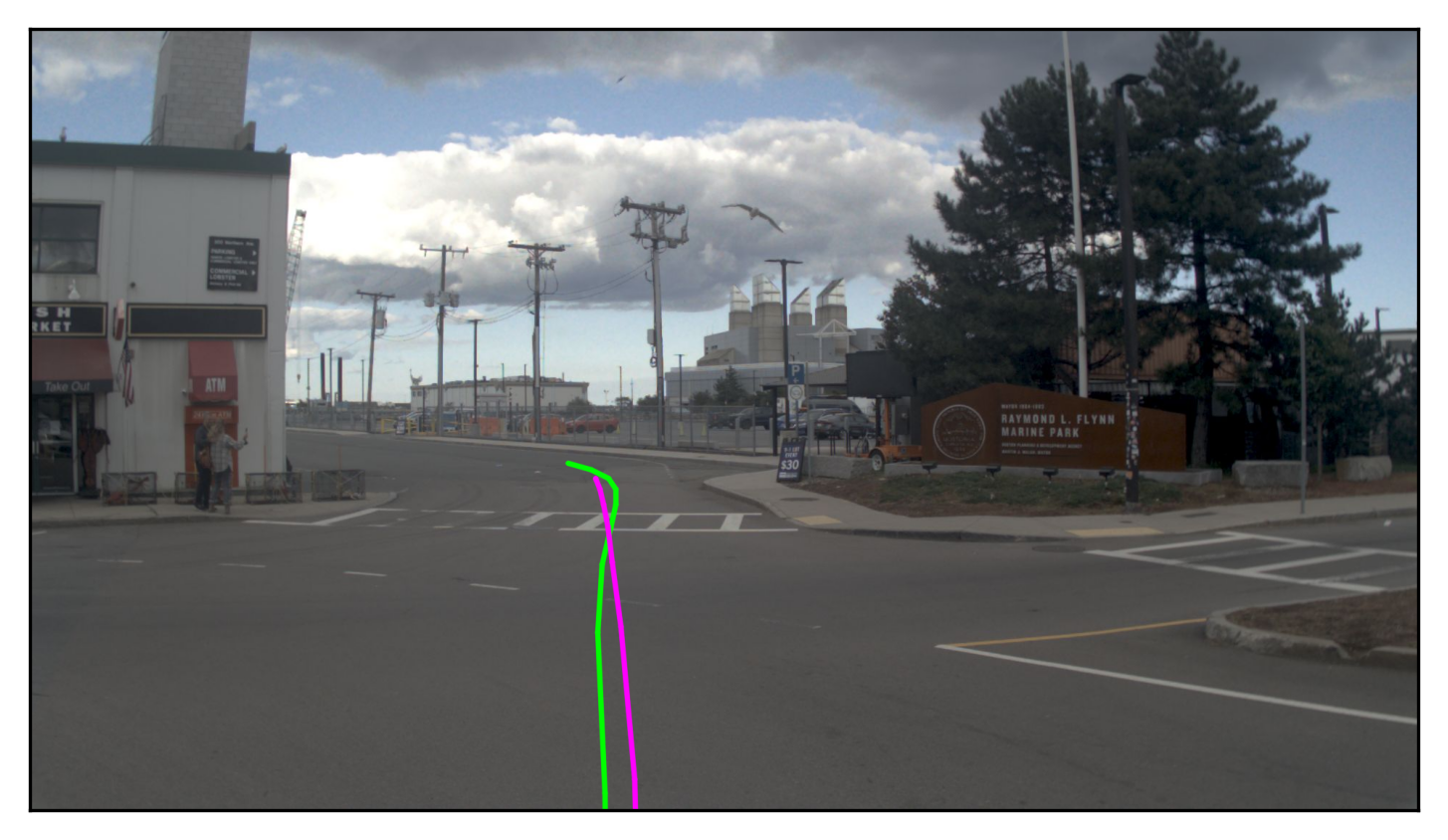}
      \hspace{-8pt}
      \includegraphics[height=0.123\linewidth]{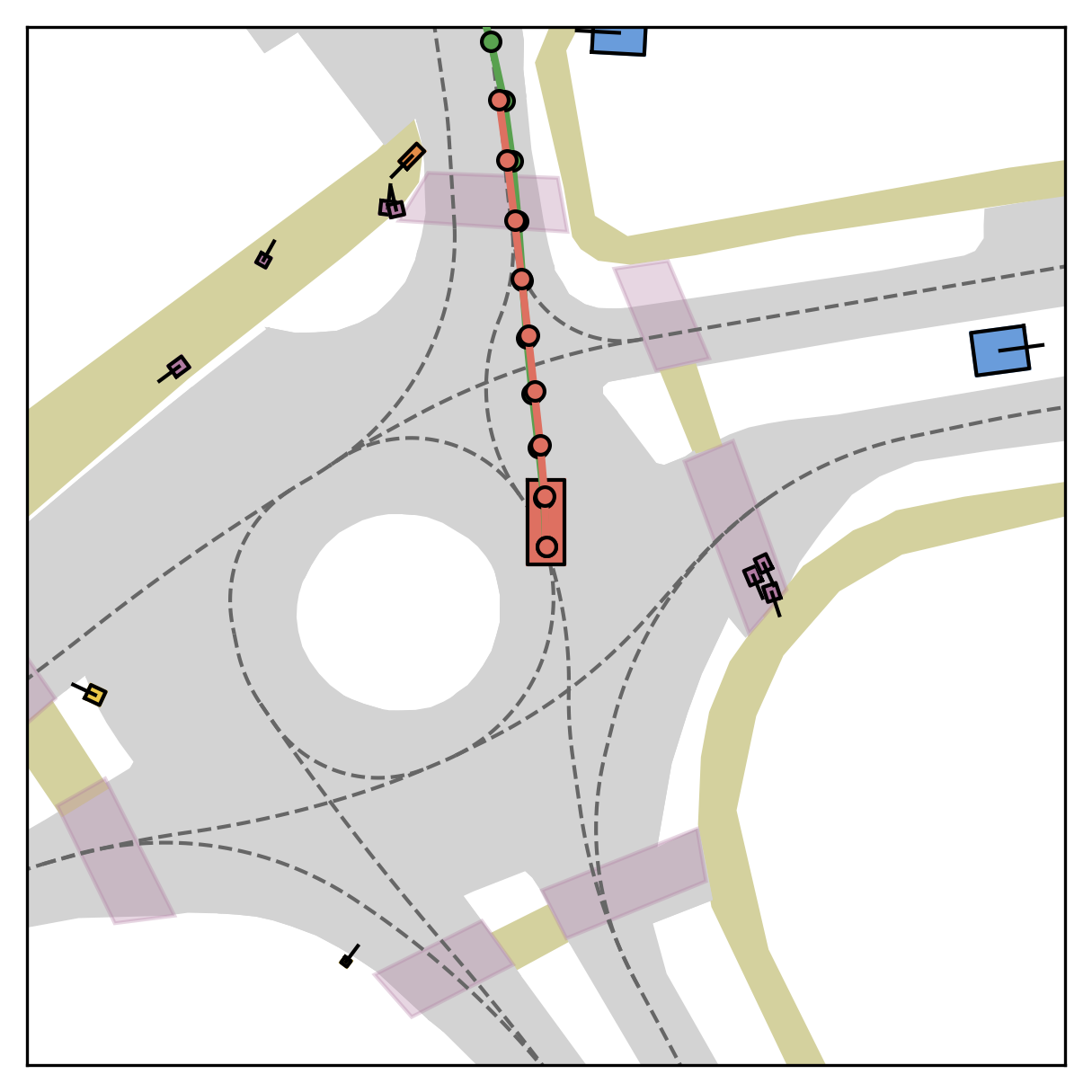}
      \caption{FlowDrive}
      \label{fig:v_b}
  \end{subfigure}
  \begin{subfigure}{\linewidth}
      \centering
      \includegraphics[width=0.5\linewidth]{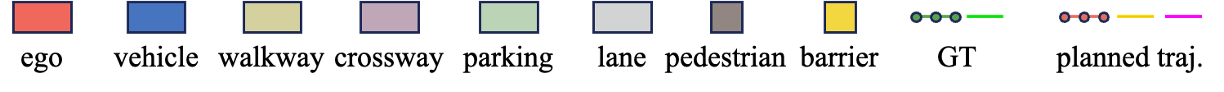}
      \vspace{-10pt}
  \end{subfigure}
  \caption{\textbf{Qualitative comparison between FlowDrive (Fig.~\ref{fig:v_a}) and the baseline DiffusionDrive (Fig.~\ref{fig:v_b}) on the navtest split.} The results indicate that FlowDrive, which incorporates energy-based flow field modeling, can generate more goal-aligned and semantically consistent trajectories compared to DiffusionDrive. It significantly improves capabilities such as drivable area compliance, lane keeping, and safety in challenging scenarios.}
  \label{fig:visual}
\end{figure*}

\subsection{Quantitative Results}
Table.~\ref{tab:quantitive} reports the quantitative comparison of FlowDrive with prior state-of-the-art methods on the Navsim benchmark. Under the image-only setting, FlowDrive with a ResNet-34 backbone achieves an EPDMS score of 84.9, surpassing DiffusionDrive (84.2) and DriveSuprim (83.1), and marking a +0.7 gain over the strongest baseline. When equipped with a stronger V2-99 backbone, FlowDrive further improves the EPDMS to 86.3, establishing a new state of the art with a +0.3 gain over the best prior baseline. When equipped with a stronger V2-99 backbone, FlowDrive further improves to an EPDMS of 86.3, setting a new state of the art and achieving a +0.3 improvement over DiffusionDrive. In addition to overall performance, FlowDrive shows notable gains in safety-related metrics such as DAC, DDC, and TTC, reflecting its ability to better respect driving constraints and avoid hazardous behaviors. At the same time, improvements in NC, EC, and LK demonstrate enhanced driving smoothness and comfort, while competitive EP scores confirm efficient goal-reaching behavior. Under the multimodal (image + LiDAR) setting, FlowDrive also achieves strong performance with both ResNet-34 and V2-99 backbones. Compared with state-of-the-art multimodal planners such as TransFuser and DiffusionDrive, FlowDrive delivers comparable or superior results, validating the effectiveness of flow-field learning and motion-decoupled diffusion planning for safe and robust trajectory generation.

\subsection{Ablation Studies}

\textbf{Ablation on different modules} \quad We conduct an ablation study to assess the impact of the three core components: Flow Field Learning, Adaptive Anchor Refinement, and Motion Decoupling, as summarized in Table.~\ref{tab:module}. Results reveal that removing any individual component leads to a noticeable drop in overall EPDMS performance. In particular, disabling the Flow Field Learning module reduces EPDMS from 86.3 to 85.8, indicating the importance of scene-level semantic and safety-aware priors for effective trajectory planning. Similarly, removing the Adaptive Anchor Refinement module results in a -0.4 EPDMS decrease, highlighting the benefit of refining trajectory anchors in a flow-aware manner for improved spatial alignment and guidance quality. Lastly, disabling the Motion Decoupling module causes a performance drop of -0.2, confirming that disentangling motion pattern prediction from trajectory generation aids task specialization and stability. 

\noindent
\textbf{Parameter Configurations of Flow Fields} \quad We further investigate how different parameter configurations affect the performance of the Flow Field module as illustrated in Table.~\ref{tab:parameter}. The results indicate that the setting with $\sigma=10.0, \eta=1.0, k_{lat}=1.0, k_{lon}=10.0$ achieves the best overall performance, reaching an EPDMS score of 86.3.

\subsection{Visualization}
Fig.~\ref{fig:visual} compares the qualitative results of FlowDrive and the baseline DiffusionDrive \cite{diffusiondrive} on the NAVSIM v2 navtest split. Leveraging physically interpretable energy-based flow field guidance, our approach produces motion plans that more accurately align with drivable areas, lane topology, and semantic intent.



\section{Conclusion}

In this work, we presented FlowDrive, a novel end-to-end autonomous driving framework that integrates flow field representations, adaptive anchor refinement, and motion-decoupled trajectory generation. By modeling spatial risk and lane priors as energy-based flow fields, FlowDrive provides interpretable and structured guidance for safe and goal-directed planning. The anchor refinement module dynamically adapts guidance points based on flow context, while our feature-level task decoupling architecture enables more specialized learning for motion intent and trajectory generation.
Extensive experiments on the NAVSIM v2 benchmark show that FlowDrive state-of-the-art performance on NAVSIM v2, outperforming prior methods not only in trajectory accuracy but also in compliance with traffic rules and dynamic scene understanding.


{
    \small
    \bibliographystyle{ieeenat_fullname}
    \bibliography{main}
}


\end{document}